\def\paperID{4179}
\def\confName{ICCV}
\def\confYear{2025}

\def\paperTitle{MGSfM: Multi-Camera Geometry Driven Global Structure-from-Motion}

\newif\ifreview 

\newif\ifarxiv 
\newcommand{\arxiv}{\arxivtrue}
\newif\ifcamera 

\newif\ifrebuttal

\arxiv

\pdfoutput=1
\documentclass[10pt,twocolumn,letterpaper]{article}
\ifreview \usepackage[review]{iccv} \fi
\ifarxiv \usepackage[pagenumbers]{iccv} \fi
\ifrebuttal \usepackage[rebuttal]{iccv} \fi
\ifcamera \usepackage{iccv} \fi

\usepackage{graphicx}	
\usepackage{amsmath}	
\usepackage{amssymb}	
\usepackage{booktabs}
\usepackage{times}
\usepackage{microtype}
\usepackage{epsfig}
\usepackage{caption}
\usepackage{float}
\usepackage{placeins}
\usepackage{color, colortbl}
\usepackage{stfloats}
\usepackage{enumitem}
\usepackage{tabularx}
\usepackage{xstring}
\usepackage{multirow}
\usepackage{xspace}
\usepackage{url}
\usepackage{bm}
\usepackage{xcolor}
\usepackage[hang,flushmargin]{footmisc}
\usepackage{adjustbox}

\ifcamera \usepackage[accsupp]{axessibility} \fi

% ...

\newcommand{\supp}{supplemental material\xspace}
\ifarxiv \renewcommand{\supp}{appendix\xspace} \fi

\newcommand{\R}[1]{{%
    \textbf{%
        \ifstrequal{#1}{1}{\textcolor{red}{mUmR}:}{%
        \ifstrequal{#1}{2}{\textcolor{blue}{UrtN}:}{%
        \ifstrequal{#1}{3}{\textcolor{teal}{HHoK}:}{%
        \ifstrequal{#1}{4}{\textcolor{cyan}{R#1}}{%
                           \textcolor{magenta}{R#1}%
        }}}}%
    }%
}}
\usepackage{xr-hyper}

\makeatletter
\newcommand*{\addFileDependency}[1]{
  \typeout{(#1)}
  \@addtofilelist{#1}
  \IfFileExists{#1}{}{\typeout{No file #1.}}
}

\makeatother

\definecolor{cvprblue}{rgb}{0.21,0.49,0.74}
\usepackage[pagebackref,breaklinks,colorlinks,allcolors=cvprblue]{hyperref}
\usepackage[capitalize]{cleveref}
\crefname{section}{Sec.}{Secs.}
\crefname{table}{Table}{Tables}
\crefname{figure}{Fig.}{Figs.}

\ifarxiv \crefname{appendix}{App.}{Apps.}
\else \crefname{appendix}{Suppl.}{Suppls.} \fi

\begin{document}
%% TITLE
\title{\paperTitle}
\author{
Peilin Tao\textsuperscript{1,2}\footnotemark[1]\quad Hainan Cui\textsuperscript{1,2}\footnotemark[1] \footnotemark[2]\quad Diantao Tu\textsuperscript{1,2}\quad Shuhan Shen\textsuperscript{1,2}\footnotemark[2]\\
\textsuperscript{1} Institute of Automation, Chinese Academy of Sciences \\
\textsuperscript{2} School of Artificial Intelligence, University of Chinese Academy of Sciences\\
{\tt\small taopeilin2023@ia.ac.cn, hncui@nlpr.ia.ac.cn, tudiantao2020@ia.ac.cn, shshen@nlpr.ia.ac.cn}
}
\maketitle
\footnotetext[1]{These authors contributed equally to this work.}

\footnotetext[2]{Corresponding author.}
\begin{abstract}

Multi-camera systems are increasingly vital in the environmental perception of autonomous vehicles and robotics. Their physical configuration offers inherent fixed relative pose constraints that benefit Structure-from-Motion (SfM). However, traditional global SfM systems struggle with robustness due to their optimization framework.
We propose a novel global motion averaging framework for multi-camera systems, featuring two core components: a decoupled rotation averaging module and a hybrid translation averaging module.
Our rotation averaging employs a hierarchical strategy by first estimating relative rotations within rigid camera units and then computing global rigid unit rotations.
To enhance the robustness of translation averaging, we incorporate both camera-to-camera and camera-to-point constraints to initialize camera positions and 3D points with a convex distance-based objective function and refine them with an unbiased non-bilinear angle-based objective function.
% Our rotation averaging employs a hierarchical strategy: first estimating relative orientations within rigid camera units, then computing global rigid unit orientations, and finally refining both.
% For translation averaging, we combine camera-to-camera and camera-to-point constraints, enhanced by an unbiased angle-based refinement method to address collinear camera motion and feature match outliers.
Experiments on large-scale datasets show that our system matches or exceeds incremental SfM accuracy while significantly improving efficiency.
Our framework outperforms existing global SfM methods, establishing itself as a robust solution for real-world multi-camera SfM applications.
The code is available at \url{https://github.com/3dv-casia/MGSfM/}.
\end{abstract}
\section{Introduction}
\label{sec:intro}
% Given a set of images, Structure-from-Motion is an effective technique for jointly recovering the camera poses and reconstructing the 3D structures. To fully perceive the environment, multi-camera systems are commonly equipped on image collection platforms and mobile robots. The physical model of the camera mount naturally provides fixed relative pose constraints for Structure-from-Motion.
% 先引出问题的重要性：多相机的普遍性，当前sfm方法在大规模场景的的问题，不鲁棒效率低，有累积误差
% 先介绍增量式的方法，再介绍全局式的方法，
% Since cameras in the multi-camera system are mounted rigidly, the internal relative poses between cameras are fixed during image collecting process. We set one camera as reference camera and the other non-reference camera poses can be derived from the reference camera poses and internal relative poses. Then, the multi-camera based SfM
% problem is formulated to only compute reference camera poses and internal relative poses. 
% To insert a figure: \input{figs/template}
% Or table: \input{tables/template}

Multi-camera systems have gained significant popularity in autonomous vehicles~\cite{heng2019project, wei2023surroundocc, liu2017robust} and mobile robotics~\cite{heng2019project, chen2021range, rubio2019review}. By rigidly mounting multiple cameras on a vehicle, these systems provide more comprehensive visual observations of the environment, enabling applications such as autonomous driving~\cite{wei2023surroundocc,liu2017robust}, visual localization and mapping~\cite{geppert2019efficient, won2020omnislam, sewtz2021robust}, and 3D object detection~\cite{han2023mmptrack, nguyen2022multi}. To reconstruct 3D structure and recover camera motions, structure-from-motion (SfM) is a fundamental technique. It begins by calculating feature correspondences between overlapping images to construct a view graph~\cite{barath2021efficient,arrigoni2023viewing}. Then, cameras are registered, and 3D points are triangulated. Finally, camera parameters and 3D points are jointly refined through bundle adjustment (BA)~\cite{triggs2000bundle,ren2022megba}. Depending on the camera registration manner, existing SfM pipelines can be classified into two categories: incremental and global methods.

Incremental SfM~\cite{schoenberger2016sfm, wu2013towards} starts with an initial pair of images to form the initial scene and iteratively register additional overlapping images via the RANSAC~\cite{chum2003locally} based Perspective-n-Point (PnP) method~\cite{gao2003complete, ding2023revisiting}. 
% After each registration, a bundle adjustment is performed to refine both camera parameters and 3D points. 
While incremental SfM is generally robust against outlier feature matches, its primary drawbacks include strong sensitivity to image registration order, time-consuming repeated bundle adjustments, and the gradual accumulation of errors as the scene expands, making it unsuitable for large-scale reconstruction tasks. 
In contrast, global SfM methods~\cite{pan2024glomap, cai2021pose} first perform rotation averaging~\cite{chatterjee2017robust, Hartley2013RotationA} to estimate global camera rotations, followed by translation averaging~\cite{ozyesil2015robust, zhuang2018baseline} to estimate global camera translations. By registering all cameras simultaneously, global approaches achieve uniform error distribution and higher efficiency. For rotation averaging, existing methods based on Lie algebra structures~\cite{chatterjee2017robust, govindu2004lie} have been extensively studied. However, translation averaging methods relying solely on relative translations encounter degeneration when the camera motion trajectory is approximately collinear~\cite{jiang2013global,manam2023sensitivity}. 
% Moreover, the normalized relative translations are error-prone in low-parallax scenes~\cite{liu2019robust,tao2024revisiting}. 
To address these issues, recent works~\cite{tao2024revisiting, pan2024glomap} jointly estimate cameras and 3D points by incorporating feature tracks into translation averaging.
% and achieve comparable accuracy and robustness than incremental methods~\cite{schoenberger2016sfm}. 

% Nevertheless, existing SfM solutions may still fail in large-scale reconstruction tasks, especially those with many mismatched feature tracks or collinear camera motions. 
Nevertheless, existing SfM solutions may still fail in large-scale reconstruction tasks, especially when confronted with numerous mismatched feature tracks or a lack of loop closures. Given that multi-camera systems are increasingly employed in image acquisition platforms, a promising approach is to exploit the inherent multi-camera constraints, whereby each rigidly mounted camera maintains a fixed internal pose relative to the other cameras within the same rigid unit.
Several works~\cite{cui2022mma, MCSfM} leverage this property by designating one camera as the reference camera, with the non-reference camera poses derived from the reference pose and the fixed internal relative poses. Thus, they reduce the SfM problem to estimating only the reference poses and the internal relative poses, leading to improved efficiency and robustness. The incremental method~\cite{MCSfM} first estimates the internal relative poses through incremental reconstruction and then incrementally registers the rigid unit to reconstruct the entire scene with fixed internal relative poses. While this approach exhibits strong robustness and mitigates scale drift with fixed internal poses, incremental registration still risks the accumulation of drift over large-scale scenes and exhibits low efficiency. The global method~\cite{cui2022mma} demonstrates extremely high efficiency by jointly estimating the reference camera poses and internal relative poses. However, jointly estimating reference camera rotations and internal relative rotations 
% via rotation averaging
is highly non-convex and sensitive to outlier relative rotations. The internal relative rotations can be estimated in advance through relative rotation averaging only when there is sufficient overlap in the field of view between two cameras in a rigid unit~\cite{cui2022mma}, which limits its broad applicability. Moreover, only camera-to-camera constraints are employed to minimize a biased distance-based objective function during translation averaging, making it less accurate to degenerate cases, such as rigid units moving along a straight line without overlapping fields of view between internal cameras.

To tackle these challenges, we introduce a novel multi-camera-geometry driven global SfM framework, abbreviated as MGSfM. By fusing multi-camera constraints, we transform the problem of estimating camera poses into the problem of estimating the rigid unit poses and internal camera poses. 
Since the rotation averaging problem is well-studied using Lie algebra structures, we decouple the multi-camera rotation averaging problem by first estimating the internal camera rotations, followed by computing the rigid unit rotations with fixed internal camera rotations. 
We begin by disregarding multi-camera constraints and applying conventional rotation averaging to estimate the internal camera rotations, thereby relaxing the overlap requirements of traditional systems. Subsequently, given the estimated internal camera rotations, the rigid unit rotation estimation employs the Baker–Campbell–Hausdorff formula~\cite{varadarajan2013lie} for linear optimization, enhancing robustness.
For the translation averaging problem, inspired by recent advancements integrating feature tracks with angle-based refinement~\cite{pan2024glomap, tao2024revisiting}, we propose a hybrid multi-camera translation averaging method. Our approach combines distance-based and angle-based objective functions while utilizing relative translations and feature tracks as hybrid inputs. This design mitigates degenerate cases and improves overall accuracy.
Camera positions are first initialized using a convex distance-based objective function under the $L_1$ norm, followed by angle-based refinement under the $L_2$ norm. Next, 3D points are triangulated using a convex cross-product-form objective function under the $L_1$ norm. Finally, with a robust initialization, a non-bilinear angle-based refinement is proposed to refine camera and 3D point positions, exhibiting improved robustness and efficiency compared to the bilinear refinement~\cite{pan2024glomap, zhuang2018baseline}.

\begin{figure}[tp]
    \setlength{\abovecaptionskip}{0.2cm}
    \centering
    \includegraphics[width=\linewidth, trim = 0mm 12mm 0mm 12mm, clip]
    {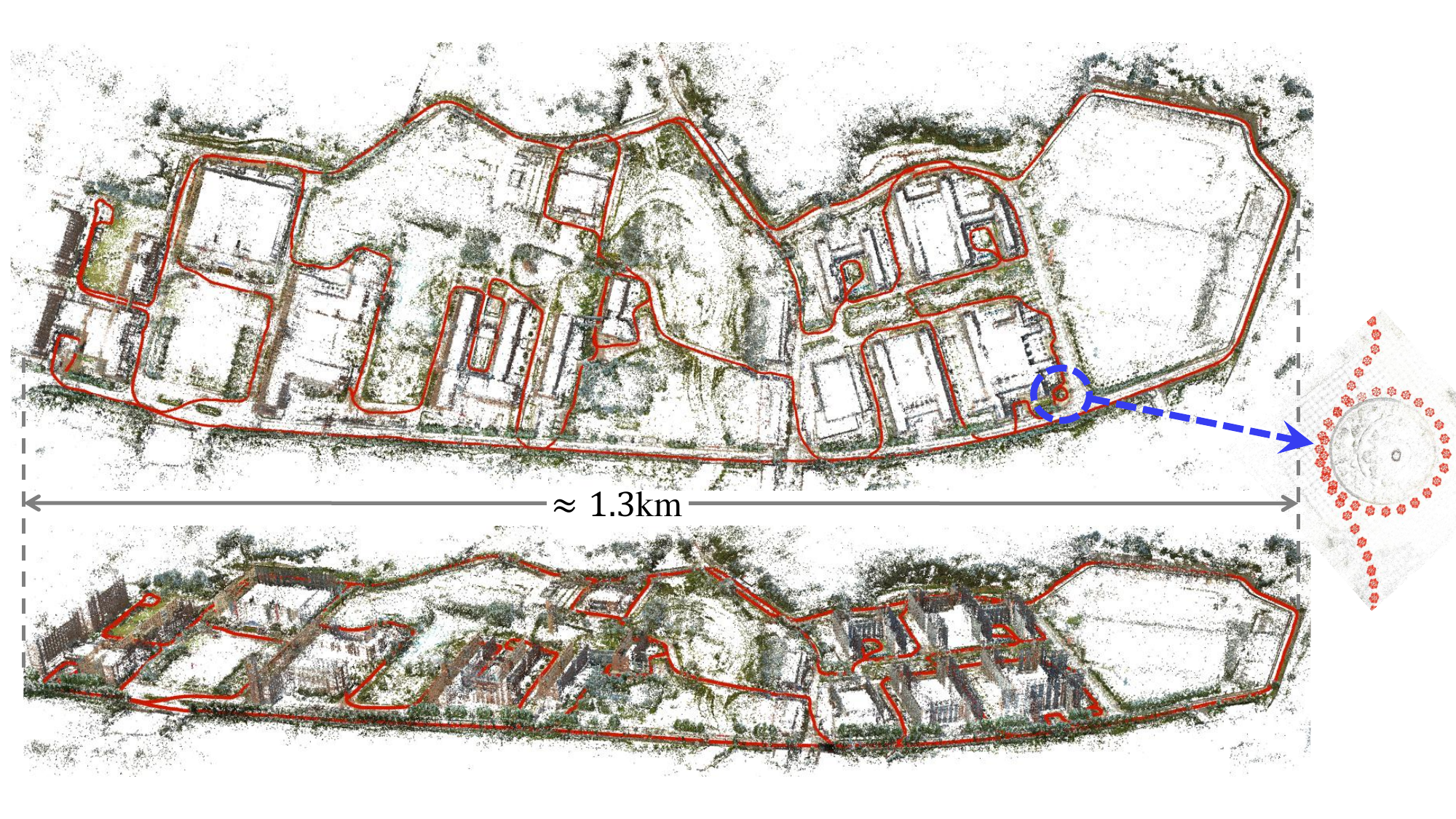}
    \caption{Our method successfully reconstructs a self-collected dataset named CAMPUS. Captured using a six-camera system, this dataset comprises over 29,000 images. Notably, many state-of-the-art methods—such as COLMAP~\cite{schoenberger2016sfm}, MCSfM~\cite{MCSfM}, GLOMAP~\cite{pan2024glomap}, and MMA~\cite{cui2022mma}—fail to produce satisfactory results, as demonstrated in \supp.
    % Our reconstruction result on a self-collected dataset captured by a six-camera system, comprising over 29,000 images, demonstrates the robustness and scalability of our system. Many state-of-the-art methods (\eg COLMAP~\cite{schoenberger2016sfm}, GLOMAP~\cite{pan2024glomap}, MCSfM~\cite{MCSfM}, and MMA~\cite{cui2022mma}) fail, as shown in \supp.
    }
    \label{fig:UCAS}
    \vspace{-0.6cm}
\end{figure}

In summary, this work makes two main \textbf{contributions}. First, we propose a decoupled multi-camera rotation averaging framework that transforms the highly non-convex multi-camera rotation averaging problem into a two-step standard single-camera problems. Second, we introduce a hybrid multi-camera translation averaging algorithm that integrates both camera-to-camera and camera-to-point constraints and incorporates a hybrid optimization framework comprising a convex distance-based initialization and an unbiased angle-based refinement. 
Various experiments show the superiority of our method compared to the state-of-the-art methods in terms of robustness, accuracy, and efficiency.
Figure~\ref{fig:UCAS} illustrates our reconstruction result on a challenging large-scale dataset captured by a six-camera Insta360 Pro2 system. While many existing methods fail to deliver satisfactory reconstructions, our approach achieves accurate results in just one hour, demonstrating its practical effectiveness in large-scale, complex real-world scenarios.
% Note that in contrast to the suggestion in GLOMAP~\cite{pan2024glomap}, which only uses feature tracks, we find that in this multi-camera-based global motion averaging, the relative translations are more vital than the feature tracks, and hybrid constraints are more effective.

% We conduct a systematic evaluation of diverse multi-camera configurations to thoroughly investigate the accuracy-efficiency trade-offs in SfM methodologies.
% Experiments on KITTI odometry~\cite{geiger2013vision}, KITTI-360~\cite{Liao2022PAMI}, and our self-collected multi-camera datasets show that our proposed method achieves accuracy comparable to or exceeding that of multi-camera incremental SfM, while significantly improving computational efficiency. Furthermore, our system outperforms state-of-the-art global motion averaging methods in both accuracy and robustness metrics.

\section{Related Work}
\label{sec:related}
\noindent\textbf{Rotation Averaging.} 
Rotation averaging is a key step in global SfM, aimed at estimating global camera rotations from pairwise relative rotations.
Early methods linearized the problem on the Lie group of rotations~\cite{govindu2004lie}. 
Crandall \etal~\cite{crandall2012sfm} initialize the camera rotations with the discrete Markov Random Field formulations and refine them with Levenberg-Marquardt optimization.
Wilson \etal~\cite{wilson2016rotations} investigate the view-graph’s density and consistency to understand when rotation averaging becomes hard. 
Chatterjee and Govindu~\cite{chatterjee2017robust} propose to initialize camera rotations via a maximum spanning tree and then refine them using an iterative re-weighted least squares (IRLS) formulation.
Dellaert \etal~\cite{dellaert2020shonan} search a sequence of higher-dimensional lifts of the rotation averaging problem to find a globally optimal solution.
Zhang \etal~\cite{zhang2023revisiting} propose to better model the underlying noise distributions by directly propagating the uncertainty from the point correspondences into the rotation averaging.
Although some incremental rotation averaging methods~\cite{gao2021incremental} are proposed, global rotation averaging methods ensure uniform error distribution for large-scale scenes.

\noindent\textbf{Translation Averaging.} 
Traditional translation averaging methods aim to derive global camera positions from pairwise relative translations.
% Quasi-convex $L_\infty$ norm-based objective functions are employed in \cite{kahl2008multiple,ke2007quasiconvex} to efficiently obtain globally optimal camera translations. However, these methods are sensitive to outliers.
Ozyesil \etal~\cite{ozyesil2015robust} propose a convex least unsquared deviations formulation to enhance the robustness.
Zhuang \etal~\cite{zhuang2018baseline} introduce an unbiased, angle-based, bilinear formulation to mitigate the impact of varying camera baseline scales.
Manam \etal~\cite{manam2022correspondence} filter outlier feature matches and refine relative translations through an iterative averaging scheme.
These methods, which rely solely on relative translations, encounter degeneration when the camera motion is collinear~\cite{jiang2013global}.
To address this problem, feature tracks are incorporated during translation averaging.
The scales of camera baselines are estimated based on the depth consistency of feature rays in~\cite{cui2015global}.
Additional constraints are formulated among cameras observing the same 3D point to eliminate the ambiguity in baseline scales in ~\cite{cui2015linear,liu2019robust,cai2021pose}.
To mitigate the impact of inaccuracies from implicitly represented 3D points, camera positions and 3D points are jointly estimated in~\cite{wilson2014robust,tao2024revisiting,pan2024glomap}.
With random initialization, a chordal distance metric angle-based objective is employed in~\cite{wilson2014robust}, and a bilinear angle-based objective is employed in~\cite{pan2024glomap}.
In contrast, camera positions and 3D points are jointly initialized using a convex objective in~\cite{tao2024revisiting}.
% Wilson \etal~\cite{wilson2014robust} formulate a chordal distance metric angle-based objective function to jointly optimize camera positions and 3D point with random initialization.
% Similarly, Pan \etal~\cite{pan2024glomap} utilize the bilinear angle-based objective function in \cite{zhuang2018baseline} to jointly optimize camera positions and 3D points with random initialization.
% Tao \etal~\cite{tao2024revisiting} select feature tracks to initialize camera positions and 3D points with convex objective function and  refine them with angle-based objective function.

\noindent\textbf{Muti-Camera Reconstruction.} 
Multi-camera-based simultaneous localization and mapping (SLAM) systems, such as ORB-SLAM2~\cite{murORB2} and ORB-SLAM3~\cite{ORBSLAM3_TRO}, leverage fixed relative poses to achieve efficient mapping. However, these systems require pre-calibrated internal camera poses and are unable to process multiple video sequences collectively, such as those obtained through crowdsourcing.
In comparison, some multi-camera-based SfM systems leverage the fixed internal relative pose constraints to estimate the relative poses automatically. 
This allows one camera to be designated the ``reference camera", reducing the parameter space to its global poses plus the constant internal relative poses. 
In incremental settings, MCSfM~\cite{MCSfM} formulates an incremental pipeline in which rigid units are successively registered, minimizing scale drift and boosting robustness thanks to the fused observations from multiple cameras. However, incremental solutions remain susceptible to gradual error accumulation and heavy repeated bundle adjustments.
On the other hand, MMA~\cite{cui2022mma} introduces a global multi-camera motion averaging framework by extending rotation and translation averaging to consider multi-camera constraints. It simultaneously solves for the reference camera poses and internal relative poses, thereby improving efficiency over incremental systems. Despite these benefits, it may fail in large-scale scenes where a large fraction of relative poses are outliers, and it cannot solve degenerate collinear configurations.
Internal camera poses in a rig are typically estimated based on a pre-built map for visual localization or calibration~\cite{heng2014Infra, hane20173d}. In contrast, our method jointly estimates both camera poses and the 3D structure in a unified optimization framework.

\section{Multi-Camera Driven Global SfM}
\label{sec:method}
In this section, we present the details of our multi-camera driven global SfM system. We first construct the view graph by computing the feature correspondences and relative poses between image pairs. 
Next, we define the multi-camera model by transforming the problem of estimating camera poses into the problem of estimating the rigid unit poses and the internal camera poses. Based on the multi-camera model, we first introduce a decoupled multi-camera rotation averaging method by estimating the internal camera rotations first, then computing the rigid unit rotations with fixed internal rotations. 
Then, we propose a hybrid, multi-camera translation averaging method that uses a hybrid distance-based and angle-based objective function with hybrid relative translations and feature tracks as input. Finally, we introduce the multi-camera bundle adjustment.
For clarity, we use calligraphic letters (e.g.,$\bm{\mathcal{C}}$,$\bm{\mathcal{R}}$,$\bm{\mathcal{P}}$) to denote the corresponding sets of variables (e.g., camera centers $\bm{c}_i$, rotations $\bm{R}_i$, 3D points $\bm{p}_k$) being optimized in the objective functions.
\subsection{View Graph Construction}
Given images collected by the multi-camera system, scale-
invariant image features~\cite{lowe2004distinctive} are first detected for all images. 
Since the multi-camera system is typically mounted on a continuously moving platform and the images are captured sequentially, we begin by sequentially matching each image with its adjacent images captured by the same camera.
Next, loop detection is performed, where each image is matched with the top-K similar images identified by the image retrieval method~\cite{schoenberger2016sfm}. 

Given initial feature matches and camera intrinsics for each image pair, we validate the feature matches using the RANSAC-based~\cite{chum2003locally}  5-point algorithm~\cite{nister2004efficient}. 
If a sufficient number of inlier feature matches are detected between image pairs, the images are considered matched. The relative rotations $\bm{R}_{ij}$ and the normalized relative translations $\bm{t}_{ij}$ are then decomposed from the corresponding essential matrices. After obtaining matched image pairs, a view graph $\mathcal{G} = \{\mathcal{V}, \mathcal{E}\}$ is constructed, where each node in $\mathcal{V}$ represents an image, and each edge $ij \in \mathcal{E}$ represents the relative pose $(\bm{R}_{ij},\bm{t}_{ij})$ and corresponding feature matches between two images $i,j \in \mathcal{V}$.
The global camera rotations and positions $(\bm{R}_{i},\bm{c}_{i})$, and the relative camera rotations and translations $(\bm{R}_{ij},\bm{t}_{ij})$ satisfy the following equations:
\begin{equation}
\label{equ:camera pose}
\bm{R}_{ij} = \bm{R}_j {\bm{R}_i}^{\top}, \quad {\bm{R}_j}^{\top}\bm{t}_{ij} = \frac{\bm{c}_i-\bm{c}_j}{||\bm{c}_i-\bm{c}_j||_2}.
\end{equation}

\subsection{Multi-Camera Model Definition}
Since the cameras in a multi-camera system are triggered approximately synchronously, the images captured simultaneously are regarded as forming a rigid unit in which the internal camera poses remain fixed.
Let $(\bm{R}_i^r,\bm{t}_i^r)$ denote the internal camera rotation and translation of image $i$ in the corresponding rigid unit coordinate system, and let $(\bm{R}_i^g,\bm{c}_i^g)$ denote the rotation and position of the rigid unit corresponding to image $i$ in the global coordinate system.
To remove rotational and positional ambiguity, we set the pose of an arbitrary internal camera and the pose of an arbitrary rigid unit to a fixed pose like $(\bm{I},\bm{0})$.
The global camera rotation and position of image $i$ can then be expressed in terms of the pose of its corresponding rigid unit and internal camera as: 
\begin{equation}
\label{equ:multi camera model}
\bm{R}_i= \bm{R}_i^r\bm{R}_i^g, \quad \bm{c}_i = \bm{c}_i^g - {\bm{R}_i}^{\top}\bm{t}_i^r.
\end{equation}
% Thus, by incorporating multi-camera constraints, global camera rotations and positions can be expressed as a combination of global rigid unit rotations and positions with internal camera rotations and translations. 
% Thus, by incorporating multi-camera constraints, this formulation enhances robustness and efficiency by increasing camera connectivity and reducing the number of unknown variables.
Thus, by incorporating multi-camera constraints, robustness and efficiency are enhanced through increased camera connectivity and a reduced number of unknown variables.

\subsection{Decoupled Multi-Camera Rotation Averaging}
Based on \cref{equ:camera pose,equ:multi camera model}, relative camera rotation $\bm{R}_{ij}$ can be represented in terms of the internal camera rotations and global rigid unit rotations as:
\begin{equation}
\label{equ:multi camera rotation}
\bm{R}_{ij}= \bm{R}_j^r\bm{R}_j^g {\bm{R}_i^g}^{\top} {\bm{R}_i^r}^{\top}.
\end{equation}
For any two images $i$ and $j$ within the same rigid unit, we have $\bm{R}_j^g = \bm{R}_i^g$.
Hence, for the relative rotations of the image pair $ij$ in the same rigid unit, \cref{equ:multi camera rotation} can be simplified as:
\begin{equation}
\label{equ:internal camera rotation}
\bm{R}_{ij}^r= \bm{R}_j^r\bm{R}_j^g {\bm{R}_i^g}^{\top} {\bm{R}_i^r}^{\top}=\bm{R}_j^r{\bm{R}_i^r}^{\top}
\end{equation}

When there is an insufficient overlapping field of view between images in the rigid units, the traditional multi-camera rotation averaging method, such as MRA~\cite{cui2022mma}, jointly estimates the global rigid unit rotations and internal camera rotations.
However, the multiplication of four unknown rotation matrices cannot use the Baker–Campbell–Hausdorff formula~\cite{varadarajan2013lie} to generate linear optimization, which renders the optimization challenging and sensitive to outlier relative rotations.
To address this, we decouple the estimation by first estimating the internal camera rotations, followed by estimating the rigid unit rotations using the known internal camera rotations.
Through decoupled estimation of internal camera rotations and global rigid unit rotations, each step can be computed effectively on Lie algebra structures~\cite{chatterjee2017robust} and use the Baker–Campbell–Hausdorff formula~\cite{varadarajan2013lie}.
\subsubsection{Internal Rotation Estimation}
Let $\rho(\cdot)$ represent the robust estimator function, and $d(\cdot)$ represents the geodesic distance metric~\cite{Hartley2013RotationA}.
We first independently estimate the global camera rotations via conventional rotation averaging by averaging all relative rotations:
\begin{equation}
\label{equ:rotation averaging}
\tilde{\bm{\mathcal{R}}} = \arg\min_{\bm{\mathcal{R}}}\sum_{i,j}\rho(d(\bm{R}_{ij},\bm{R}_j\bm{R}_i^{\top})).
\end{equation}
We employ the method of Chatterjee \etal~\cite{chatterjee2017robust} to solve \cref{equ:rotation averaging}. Specifically, camera rotations are first initialized using image pairs from the maximum spanning tree of the view graph. Then, they are refined under the $L_1$ norm for a few iterations, followed by further refinement using the IRLS method.
% based on the number of inlier feature matches for each matched image pair,

% Then, we solve \cref{equ:rotation averaging} under $L_1$ norm for a few iterations, followed by refining the solution through the IRLS method until convergence.

Given the known initial global camera rotations $\tilde{\bm{\mathcal{R}}}$ from \cref{equ:rotation averaging}, the relative camera rotations of image pairs in the same rigid unit from \cref{equ:internal camera rotation} are calculated by:
\begin{equation}
\label{equ:new relative rotation}
\bm{R}_j^r{\bm{R}_i^r}^{\top} = \bm{R}_{ij}^r = \tilde{\bm{R}}_j {\tilde{\bm{R}}_i}^{\top}.
\end{equation}
To remove rotational ambiguity, we assume that the internal rotation $\bm{R}_i^r$ in each rigid unit is set to the identity matrix.
From \cref{equ:new relative rotation}, the remaining internal rotations for each image like $j$ in the same rigid unit are computed as: 
\begin{equation}
\hat{\bm{R}}_j^r = \tilde{\bm{R}}_j {\tilde{\bm{R}}_i}^{\top}\bm{R}_i^r= \tilde{\bm{R}}_j {\tilde{\bm{R}}_i}^{\top}.
% \tilde{\bm{R}}_j^r = \tilde{\bm{R}}_j {\tilde{\bm{R}}_i}^{\top}.
\end{equation}
Since each rigid unit provides a solution for each remaining internal camera rotation, we obtain a set of possible solutions for each remaining internal camera rotation. To enhance robustness against outliers, we employ the method in \cite{hartley2011l1} to estimate each internal camera rotation $\bm{R}^r$ by computing the geodesic median of all potential estimates $\hat{\bm{\mathcal{R}}^r}$ in $\mathrm{SO}(3)$.
 % through a two-step process: first, are transformed into their angle-axis representations, followed by computing the median of these angle-axis vectors to obtain the final estimation.
% The objective function is formulated as follows: 
% \begin{equation}
% \label{equ:internal rotation averaging}
% \min_{\bm{R}^r}\sum_{ij}\rho(d(\tilde{\bm{R}}_j {\tilde{\bm{R}}_i}^{\top}, \bm{R}_j^r{\bm{R}_i^r}^{\top})).
% \end{equation}
\subsubsection{Rigid Unit Rotation Estimation}
With the known internal camera rotations and given all relative camera rotations in the view graph, the relative rigid unit rotations are computed by transforming \cref{equ:multi camera rotation}: 
\begin{equation}
\bm{R}_{ij}^g = \bm{R}_j^g {\bm{R}_i^g}^{\top} = {\bm{R}_j^r}^{\top}\bm{R}_{ij}\bm{R}_i^r .
\end{equation}
Then, global rigid unit rotations are estimated through rotation averaging on Lie algebra structures. Similar to \cref{equ:rotation averaging}, the objective function is formulated by:
\begin{equation}
\label{equ:multi rotation averaging}
\min_{\bm{\mathcal{R}}^g}\sum_{i,j}\rho(d(\bm{R}_{ij}^g,\bm{R}_j^g {\bm{R}_i^g}^{\top} )),
\end{equation}
which can also be effectively solved with the method of Chatterjee \etal~\cite{chatterjee2017robust}.
To leverage the larger field of view provided by multiple cameras for enhanced robustness, we construct a maximum spanning tree based on the number of inlier feature matches from all overlapping image pairs between rigid units, and use it to estimate the initial global rotations of the rigid units.
Finally, given the estimated internal camera rotations ${\bm{R}_j^r}$ and global rigid unit rotations ${\bm{R}_j^g}$, we obtain the global camera rotations using \cref{equ:multi camera model}.

% , and subsequently filter out relative rotation outliers with angular errors exceeding $\alpha$ based on these initial rotations.
% $\alpha$ is set to $15^\circ$ in our paper.

% In conclusion, by decoupling the estimation of internal camera rotations from that of global rigid unit rotations, we leverage the effectiveness of existing rotation averaging methods based on Lie algebra structures. These methods exhibit strong robustness and do not require cameras in a multi-camera system to have overlapping fields of view.
In conclusion, our approach decouples the estimation of internal camera rotations from global rigid unit rotations without requiring cameras in a multi-camera system to have overlapping fields of view. Leveraging the effectiveness of existing methods based on Lie algebra structures, our method demonstrates strong robustness and efficiency.
% compared to the joint estimation method MMA~\cite{cui2022mma}.

\subsection{Hybrid Multi-Camera Translation Averaging}
From \cref{equ:camera pose,equ:multi camera model}, relative translations $\bm{t}_{ij}$ are expressed in terms of global rigid unit positions and internal camera translations as follows:
\begin{align}
\label{equ:multi camera position}
{\bm{R}_j}^{\top}\bm{t}_{ij} &= \frac{\bm{\mathbb{C}}_{ij}}{||\bm{\mathbb{C}}_{ij}||_2}, \\
\text{where} \; \bm{\mathbb{C}}_{ij} \triangleq \bm{c}_{i} - \bm{c}_{j} & = \bm{c}_i^g - \bm{c}_j^g + {\bm{R}_j}^{\top}\bm{t}_j^r - {\bm{R}_i}^{\top}\bm{t}_i^r.\notag
\end{align}

Existing work like MMA~\cite{cui2022mma} performs translation averaging by employing only relative translations to formulate a convex distance-based objective function by: 
\begin{equation}
\label{equ:distance-based MTA}
\min_{\substack{\bm{\mathcal{C}}^g,\mathcal{S},\bm{\mathcal{T}}^r}}
\sum_{i,j}||s_{ij}{\bm{R}_j}^{\top}\bm{t}_{ij} - \bm{\mathbb{C}}_{ij}||_1, \;
\mathrm{s.t.} \;  s_{ij}\ge 1.
\end{equation}

% \begin{equation}
% \label{equ:distance-based MTA}
% \min_{\substack{\bm{\mathcal{C}},\mathcal{S}}}
% \sum_{i,j}||s_{ij}{\bm{R}_j}^{\top}\bm{t}_{ij} - (\bm{c}_{i}-\bm{c}_{j})||_1, \;
% \mathrm{s.t.} \;  s_{ij}\ge 1.
% \end{equation}

Here $s_{ij}$ denotes the scale between camera $i$ and camera $j$. 
Although \cref{equ:distance-based MTA} is a convex problem, the biased distance-based objective function compromises accuracy~\cite{zhuang2018baseline}.
Two types of angle-based objectives are usually used to address this.
A non-bilinear objective function is formulated as:
\begin{equation}
\label{equ:non-bilinear}
\min_{\substack{\bm{\mathcal{C}}}}\sum_{i,j} \rho(||{\bm{R}_j}^{\top}\bm{t}_{ij} - \frac{\bm{c}_i - \bm{c}_j}{||\bm{c}_i - \bm{c}_j||_2}||_2).
\end{equation}

% \begin{equation}
% \label{equ:tmp-non-bilinear}
% \min_{\substack{\bm{\mathcal{C}}, \bm{\mathcal{P}}}}\sum_{i,j} \rho(||{\bm{R}_i}^{\top}\bm{f}_{ik} - \frac{\bm{p}_k - \bm{c}_i}{||\bm{p}_k - \bm{c}_i||_2}||_2).
% \end{equation}

While efficient, such a formulation with random initialization is sensitive to outliers~\cite{wilson2014robust}.
% While efficient, such a formulation needs a good initialization and may be sensitive to outliers, as shown in~\cite{wilson2014robust}.
A bilinear objective function is formulated by incorporating normalizing variables $d_{ij}$ as:
\begin{equation}
\label{equ:bilinear}
\min_{\substack{\bm{\mathcal{C}},\mathcal{D}}}\sum_{i,j} \rho(||{\bm{R}_j}^{\top}\bm{t}_{ij} - {d_{ij}}(\bm{c}_i - \bm{c}_j)||_2), \; \mathrm{s.t.} \;  d_{ij}\ge 0.
\end{equation}
While incorporating additional variables improves convergence, it compromises efficiency. Furthermore, the exclusive use of camera-to-camera constraints derived from relative translations renders these methods susceptible to degenerate cases~\cite{tao2024revisiting, jiang2013global}. This issue can be addressed by incorporating camera-to-point constraints. The mathematical expression of the camera-to-point constraint is given by:
\begin{equation}
\label{equ:camera-to-point}
{\bm{R}_i}^{\top}\bm{f}_{ik} = \frac{\bm{p}_k-\bm{c}_i}{||\bm{p}_k-\bm{c}_i||_2}, \quad\bm{f}_{ik} = \frac{\pi_i^{-1}(\bm{x}_{ik})}{||\pi_i^{-1}(\bm{x}_{ik})||_2},  
\end{equation}
where $\bm{p}_k$ is the position of point k, $\bm{x}_{ik}$ is the feature point in image $i$ corresponding to point $k$,  $\pi_i$ is the intrinsics of camera $i$ and $\bm{f}_{ik}$ is the normalized feature ray from image $i$ to point $k$ in the camera coordinate system.
The method GLOMAP~\cite{pan2024glomap} exclusively uses feature tracks to formulate a bilinear angle-based objective function as follows:
\begin{equation}
\label{equ:glomap}
\min_{\substack{\bm{\mathcal{C}},\mathcal{D},\bm{\mathcal{P}}}}\sum_{i,k} \rho(||{\bm{R}_i}^{\top}\bm{f}_{ik}-d_{ik}(\bm{p}_k-\bm{c}_i)
||_2),\; \mathrm{s.t.\,}\; d_{ik}\ge0.
\end{equation}
% Based on \cref{equ:glomap}, all camera positions and 3D points are jointly estimated using random initialization. 
% Although this function converges fast, the random initialization in~\cite{pan2024glomap} for this non-convex problem undermines robustness, and redundant variables make the computation time suffer.
Although this bilinear formulation converges reliably~\cite{pan2024glomap}, the random initialization strategy used to solve this non-convex problem compromises robustness, and inclusion of redundant variables increase computational overhead.
% To enhance robustness and efficiency, 
% we draw inspiration from \cite{tao2024revisiting} which provides initial camera positions and 3D points for the followed angle-based refinement.
% We first employ the method in \cite{tao2024revisiting} to re-estimate the relative translations with known camera rotations.
% Given re-estimated relative translations, 

In summary, a single optimization framework usually cannot fulfill the large-scale translation averaging task. Hence, we propose to handle this in a hybrid way, where the distance-based method is used for robust initialization, and the non-bilinear angle-based method is used for further refinement.

\subsubsection{Camera Position Initialization}
Given global camera rotations and relative translations from the view graph, camera positions are first initialized using \cref{equ:distance-based MTA} and subsequently refined using a non-bilinear objective function, as formulated below:
% rather than the bilinear formulation employed in \cite{pan2024glomap, zhuang2018baseline} to achieve improved performance
% First, we use the \cref{equ:distance-based MTA} to estimate initial camera positions. In contrast to \cite{pan2024glomap, zhuang2018baseline}, which performs refinement using a bilinear objective function, we refine the initial camera positions using a non-bilinear objective function to achieve improved performance, formulated as follows:
\begin{equation}
\label{equ:non-bilinear trans angle-based}
\min_{\substack{\bm{\mathcal{C}}^g,\bm{\mathcal{T}}^r}}\sum_{i,j} \rho(||{\bm{R}_j}^{\top}\bm{t}_{ij} - \frac{\bm{\mathbb{C}}_{ij}}{||\bm{\mathbb{C}}_{ij}||_2}||_2).
\end{equation}
\subsubsection{Robust 3D Point Triangulation}
With estimated global camera poses, we utilize the convex distance-based cross-product-form objective function to robustly triangulate the initial 3D points under $L_1$ norm as:
\begin{equation}
\label{eq:Triangulation}
\min_{\substack{\bm{\mathcal{P}}}}
\sum_{i,k}||{\bm{R}_i}^{\top}\bm{f}_{ik}\times(\bm{p}_{k}-\bm{c}_{i})||_1.
\end{equation}
\subsubsection{Joint Angle-based Refinement}
Finally, based on \cref{equ:camera-to-point,equ:multi camera model}, we derive the camera-to-point constraints from feature tracks to jointly refine all camera positions and 3D points through the multi-camera-based non-bilinear angle-based objective function as: 
\begin{equation}
\label{equ:angle-based all}
\min_{\substack{\bm{\mathcal{C}}^g, \bm{\mathcal{P}},\bm{\mathcal{T}}^r}}\sum_{i,k} \rho(||{\bm{R}_i}^{\top}\bm{f}_{ik} -\frac{\bm{p}_k - \bm{c}_i^g + {\bm{R}_i}^{\top}\bm{t}_i^r}{||\bm{p}_k - \bm{c}_i^g + {\bm{R}_i}^{\top}\bm{t}_i^r||_2}||_2).
\end{equation}

In conclusion, relative translations and feature tracks within a robust distance-based objective function primarily serve as initialization, while feature tracks within a non-bilinear objective function are mainly utilized for refinement.
% By providing a reasonable initialization of camera positions and 3D points, non-bilinear angle-based refinement achieves improved robustness and efficiency compared to the bilinear formulation, as demonstrated in \cref{sec:experiment}.

\subsection{Multi-Camera Bundle Adjustment}
Given camera parameters and 3D points, conventional bundle adjustment refines these parameters simultaneously by minimizing the reprojection error, which is formulated as:
\begin{equation}
\label{eq:ba}
\min_{\substack{\bm{\mathcal{C}},\bm{\mathcal{P}},\bm{\mathcal{R}}, \Pi}}
\sum_{i,k}\rho(||\pi_i(\bm{R}_i(\bm{p}_k-\bm{c}_i ))-\bm{x}_{ik}||_2).
\end{equation}
By integrating multi-camera constraints from \cref{equ:multi camera model}, the multi-camera bundle adjustment is formulated as:
\begin{equation}
\label{eq:mba}
\min_{\substack{\bm{\mathcal{C}}^g,\bm{\mathcal{P}},\bm{\mathcal{R}}^g ,\\\bm{\mathcal{R}}^r,\bm{\mathcal{T}}^r,\Pi}}
\sum_{i,k}\rho(||\pi_i(\bm{R}_i^r\bm{R}_i^g(\bm{p}_k-\bm{c}_i^g )+\bm{t}_i^r)-\bm{x}_{ik}||_2).
\end{equation}
% Similar to \cite{pan2024glomap,moulon2016openmvg}, 
% In our system, the rotations of internal cameras and rigid units are initially fixed in the first round of multi-camera bundle adjustment and subsequently optimized with the 3D points and translations. 
% Finally, to address the problem of possible asynchronous triggering, all camera poses are independently refined using bundle adjustment described in \cref{eq:ba}.
In our framework, the rotations of internal cameras and rigid units remain fixed during the initial round of multi-camera bundle adjustment and are subsequently jointly optimized alongside 3D points and translations. 
% To further address potential asynchronous triggering issues, all camera poses are independently optimized through a final bundle adjustment process detailed in \cref{eq:ba}.
\begin{table*}[t]
\setlength{\tabcolsep}{2.3pt}
\centering
\begin{tabular}{cc|cccc|cccc|cccc|cccc|cccc|cccc}
\hline
\multicolumn{2}{c|}{Data}        & \multicolumn{4}{c|}{COLMAP\cite{schoenberger2016sfm}}           & \multicolumn{4}{c|}{GLOMAP\cite{pan2024glomap}}                 & \multicolumn{4}{c|}{MMA\cite{cui2022mma}}                     & \multicolumn{4}{c|}{MCSfM\cite{MCSfM}}                          & \multicolumn{4}{c|}{DMRA+MGP}                                                     & \multicolumn{4}{c}{MGSfM}                                                        \\ \hline
\multicolumn{1}{c|}{Name} & $N$  & $\tilde{e}_r$ & \multicolumn{1}{c|}{$\bar{e}_r$}  & $\tilde{e}_t$ & $\bar{e}_t$  & $\tilde{e}_r$ & \multicolumn{1}{c|}{$\bar{e}_r$}  & $\tilde{e}_t$ & $\bar{e}_t$  & $\tilde{e}_r$ & \multicolumn{1}{c|}{$\bar{e}_r$} & $\tilde{e}_t$ & $\bar{e}_t$ & $\tilde{e}_r$ & \multicolumn{1}{c|}{$\bar{e}_r$}  & $\tilde{e}_t$ & $\bar{e}_t$  & $\tilde{e}_r$ & \multicolumn{1}{c|}{$\bar{e}_r$}  & $\tilde{e}_t$ & $\bar{e}_t$  & $\tilde{e}_r$ & \multicolumn{1}{c|}{$\bar{e}_r$}  & $\tilde{e}_t$ & $\bar{e}_t$  \\ \hline
\multicolumn{1}{c|}{00}   & 9082 & 0.4           & \multicolumn{1}{c|}{0.5}          & 0.9           & 3.8          & 0.4           & \multicolumn{1}{c|}{\textbf{0.4}} & 0.8           & 1.3          & 0.7           & \multicolumn{1}{c|}{0.8}         & 1.5           & 1.9         & \textbf{0.3}  & \multicolumn{1}{c|}{\textbf{0.4}} & 0.6           & 0.8          & 0.4           & \multicolumn{1}{c|}{\textbf{0.4}} & \textbf{0.5}  & \textbf{0.7} & 0.4           & \multicolumn{1}{c|}{\textbf{0.4}} & \textbf{0.5}  & \textbf{0.7} \\
\multicolumn{1}{c|}{01}   & 2202 & \textbf{0.2}  & \multicolumn{1}{c|}{0.5}          & 1.5           & 3.0          & 0.5           & \multicolumn{1}{c|}{1.4}          & 4.5           & 1e2          & 0.7           & \multicolumn{1}{c|}{1.1}         & 16.3          & 23.2        & \textbf{0.2}  & \multicolumn{1}{c|}{0.7}          & 1.2           & 2.7          & 0.4           & \multicolumn{1}{c|}{1.4}          & 0.8           & 31.7         & \textbf{0.2}  & \multicolumn{1}{c|}{\textbf{0.3}} & \textbf{0.6}  & \textbf{1.1} \\
\multicolumn{1}{c|}{02}   & 9322 & 0.5           & \multicolumn{1}{c|}{0.7}          & 4.0           & 12.3         & 0.4           & \multicolumn{1}{c|}{0.5}          & 5.4           & 1e3          & 1.8           & \multicolumn{1}{c|}{2.3}         & 7.0           & 10.1        & \textbf{0.3}  & \multicolumn{1}{c|}{\textbf{0.3}} & 1.0           & \textbf{1.4}          & 0.4           & \multicolumn{1}{c|}{0.4}          & 1.3           & 1.8          & 0.4           & \multicolumn{1}{c|}{0.4}          & \textbf{0.9}  & \textbf{1.4} \\
\multicolumn{1}{c|}{03}   & 1602 & 0.1           & \multicolumn{1}{c|}{0.2}          & 0.2           & 0.8          & 0.2           & \multicolumn{1}{c|}{0.2}          & 0.4           & 0.9          & 0.7           & \multicolumn{1}{c|}{0.7}         & 0.4           & 1.9         & \textbf{0.2}  & \multicolumn{1}{c|}{\textbf{0.2}} & \textbf{0.2}  & \textbf{0.4} & \textbf{0.2}  & \multicolumn{1}{c|}{\textbf{0.2}} & \textbf{0.2}  & \textbf{0.4} & \textbf{0.2}  & \multicolumn{1}{c|}{\textbf{0.2}} & \textbf{0.2}  & \textbf{0.4} \\
\multicolumn{1}{c|}{04}   & 542  & \textbf{0.1}  & \multicolumn{1}{c|}{\textbf{0.1}} & \textbf{0.1}  & \textbf{0.2} & \textbf{0.1}  & \multicolumn{1}{c|}{\textbf{0.1}} & 0.2           & 0.5          & 0.3           & \multicolumn{1}{c|}{0.3}         & 2.4           & 13.3        & \textbf{0.1}  & \multicolumn{1}{c|}{0.2}          & \textbf{0.1}  & \textbf{0.2} & \textbf{0.1}  & \multicolumn{1}{c|}{\textbf{0.1}} & \textbf{0.1}  & \textbf{0.2} & \textbf{0.1}  & \multicolumn{1}{c|}{\textbf{0.1}} & \textbf{0.1}  & \textbf{0.2} \\
\multicolumn{1}{c|}{05}   & 5522 & 0.6           & \multicolumn{1}{c|}{0.6}          & 0.8           & 2.2          & \textbf{0.2}  & \multicolumn{1}{c|}{\textbf{0.3}} & 0.3           & 0.6          & 0.6           & \multicolumn{1}{c|}{1.3}         & 0.6           & 1.2         & 0.3           & \multicolumn{1}{c|}{\textbf{0.3}} & \textbf{0.2}  & 0.5          & \textbf{0.2}  & \multicolumn{1}{c|}{\textbf{0.3}} & \textbf{0.2}  & \textbf{0.3} & \textbf{0.2}  & \multicolumn{1}{c|}{\textbf{0.3}} & \textbf{0.2}  & \textbf{0.3} \\
\multicolumn{1}{c|}{06}   & 2202 & 0.2           & \multicolumn{1}{c|}{0.4}          & 0.2           & 1.6          & \textbf{0.1}  & \multicolumn{1}{c|}{\textbf{0.2}} & 0.3           & 0.5          & 0.5           & \multicolumn{1}{c|}{0.5}         & 0.3           & 1.0         & 0.2           & \multicolumn{1}{c|}{0.4}          & \textbf{0.1}  & 0.8          & \textbf{0.1}  & \multicolumn{1}{c|}{\textbf{0.2}} & \textbf{0.1}  & \textbf{0.1} & \textbf{0.1}  & \multicolumn{1}{c|}{\textbf{0.2}} & \textbf{0.1}  & \textbf{0.1} \\
\multicolumn{1}{c|}{07}   & 2202 & 1.1           & \multicolumn{1}{c|}{1.1}          & 0.5           & 2.2          & 0.3           & \multicolumn{1}{c|}{\textbf{0.3}} & 0.3           & \textbf{0.3} & 0.6           & \multicolumn{1}{c|}{0.6}         & 0.9           & 2.2         & 0.5           & \multicolumn{1}{c|}{0.7}          & 0.7           & 3.1          & \textbf{0.2}  & \multicolumn{1}{c|}{\textbf{0.3}} & \textbf{0.2}  & \textbf{0.3} & \textbf{0.2}  & \multicolumn{1}{c|}{\textbf{0.3}} & \textbf{0.2}  & \textbf{0.3} \\
\multicolumn{1}{c|}{08}   & 8142 & 0.7           & \multicolumn{1}{c|}{0.9}          & 6.1           & 11.5         & 0.7           & \multicolumn{1}{c|}{0.9}          & 3.1           & 4.7          & 0.8           & \multicolumn{1}{c|}{0.8}         & 2.6           & 3.1         & \textbf{0.4}  & \multicolumn{1}{c|}{0.6}          & 1.2           & 2.6          & \textbf{0.4}  & \multicolumn{1}{c|}{0.5}          & 1.0           & 1.5          & \textbf{0.4}  & \multicolumn{1}{c|}{\textbf{0.4}} & \textbf{0.9}  & \textbf{1.4} \\
\multicolumn{1}{c|}{09}   & 3182 & 0.6           & \multicolumn{1}{c|}{0.6}          & 1.1           & 3.4          & \textbf{0.3}  & \multicolumn{1}{c|}{\textbf{0.3}} & 1.7           & 2.8          & 0.6           & \multicolumn{1}{c|}{0.6}         & 1.7           & 3.5         & \textbf{0.3}  & \multicolumn{1}{c|}{\textbf{0.3}} & \textbf{0.5}  & \textbf{1.0} & \textbf{0.3}  & \multicolumn{1}{c|}{\textbf{0.3}} & \textbf{0.5}  & \textbf{1.0} & \textbf{0.3}  & \multicolumn{1}{c|}{\textbf{0.3}} & \textbf{0.5}  & \textbf{1.0} \\
\multicolumn{1}{c|}{10}   & 2402 & 0.8           & \multicolumn{1}{c|}{1.4}          & 2.5           & 4.3          & \textbf{0.3}  & \multicolumn{1}{c|}{\textbf{0.3}} & 0.7           & 1.3          & 0.7           & \multicolumn{1}{c|}{0.8}         & 0.6           & 1.2         & 0.4           & \multicolumn{1}{c|}{0.5}          & \textbf{0.4}  & \textbf{1.2} & 0.4           & \multicolumn{1}{c|}{0.6}          & 0.6           & 1.4          & 0.4           & \multicolumn{1}{c|}{0.5}          & 0.6           & 1.3          \\ \hline
\end{tabular}
\caption{Camera pose accuracy on the KITTI Odometry dataset. $N$ shows the number of images; $\tilde{e}_r$ and $\bar{e}_r$ respectively denote the median and mean rotation error in degrees; $\tilde{e}_t$ and $\bar{e}_t$ respectively denote the median and mean position error in meters. The best results are \textbf{bolded}.} 	
\vspace{-0.3cm}
\label{tab:KITTI}
\end{table*}
\section{Experiment}
\label{sec:experiment}
% The experiments were performed on an Ubuntu 20.04.5 LTS platform equipped with 64GB of memory and a 12th Gen Intel(R) Core(TM) i7-12700 CPU @ 2.10GHz with 20 cores. 
We conduct experiments on real images collected by various multi-camera systems, including a stereo camera from the KITTI Odometry dataset~\cite{geiger2013vision}, a four-camera system from the KITTI-360 dataset~\cite{Liao2022PAMI}, two types of multi-camera systems mounted on street-view cars, and the Insta360 Pro2 system equipped with six fisheye cameras. Notably, only the two KITTI datasets provide ground-truth camera poses. 

% The details of the testing datasets are shown in \cref{tab:camera_systems}.

We compare our system MGSfM, with four state-of-the-art SfM techniques: the incremental method COLMAP~\cite{schoenberger2016sfm}, the global method GLOMAP~\cite{pan2024glomap}, the multi-camera-based incremental method MCSfM~\cite{MCSfM}, and the multi-camera-based global method MMA~\cite{cui2022mma}.
Additionally, we propose a method named ``DMRA+MGP", which employs our decoupled multi-camera rotation averaging (DMRA) and multi-camera bundle adjustment techniques, but modifies the hybrid multi-camera translation averaging step by integrating multi-camera constraints from \cref{equ:multi camera model} into the global positioning (MGP) objective function of GLOMAP~\cite{pan2024glomap} (see \cref{equ:glomap}). For the DMRA step, we utilize the method suggested in \cite{chatterjee2017robust} to solve \cref{equ:rotation averaging,equ:multi rotation averaging}. 
We employ the alternating direction method of multipliers (ADMM) optimizer~\cite{boyd2011distributed} to solve the $L_1$ norm optimization problem in \cref{equ:distance-based MTA,eq:Triangulation}. We use the Levenberg-Marquardt algorithm~\cite{levenberg1944method} from Ceres~\cite{Agarwal_Ceres_Solver_2022} as the optimizer to solve both the angle-based refinement problem and the bundle adjustment.
In this framework, the Cauchy kernel is employed in \cref{equ:non-bilinear trans angle-based,equ:angle-based all}, while the Huber kernel is applied in \cref{eq:mba}.

All methods share the same view graph as input, which contains relative poses and the corresponding feature matches. For accuracy evaluation, we compare the median and mean errors of both camera rotations and positions after bundle adjustment on two kinds of KITTI datasets. We compare the reconstruction runtime on the larger KITTI-360 dataset for efficiency evaluation, excluding the view graph construction step since it is identical across all methods.

\begin{figure}[tp]
	\setlength{\belowcaptionskip}{-0.1cm}
    \centering
    \includegraphics[width=\linewidth, trim = 0mm 50mm 0mm 50mm, clip]{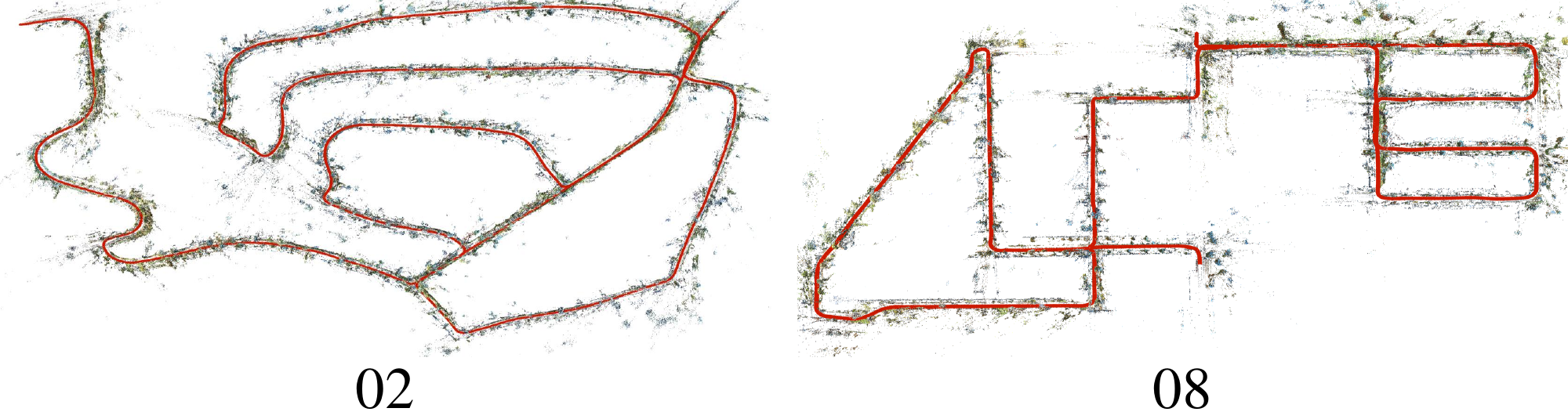}
    \caption{Our reconstruction results for data 02 and data 08 of the KITTI Odometry datasets.}
    \label{fig:kitti}
    \vspace{-0.1cm}
\end{figure}

\subsection{Evaluation of KITTI Odometry Datasets}
This dataset was collected using a stereo camera mounted on a driving car, where most camera motion trajectories are approximately collinear. Since there are moving cars and pedestrians on the street, we first use an off‐the‐shelf semantic segmentation algorithm~\cite{chen2017rethinking} to detect these objects and avoid extracting feature points from these regions. 

The accuracy comparison of the camera poses estimated by different methods is presented in \cref{tab:KITTI}, with MGSfM generally achieving the highest accuracy among most datasets. For scenes without loop closure, such as data 02 and 08, classical SfM methods yield camera poses with low accuracy due to scale drift. 
Comparing the COLMAP and MCSfM, the incorporation of multi-camera constraints improves the accuracy of the estimated camera poses, but the accumulated error still causes scene drift. 
In comparison, our global multi-camera system is more accurate.
Since there is an overlap between the stereo cameras, global rotations of reference cameras in MMA~\cite{cui2022mma} are estimated with known internal camera rotations. 
% However, it still exhibits large position estimation errors since only relative translations are used. 
However, it still exhibits large position errors because only a biased distance-based objective function is used.
For data 01, which contains numerous outlier feature matches, MGSfM also yields the best camera poses, while DMRA+MGP fails, demonstrating the strong robustness of our hybrid translation averaging.
\cref{fig:kitti} presents two sample reconstruction results on the large-scale data 02 and 08, which are produced by our system MGSfM.

% 时间的图，kitti和kitti360，横轴为各个数据集，按照图像数目从小到大排列，纵轴为时间
\begin{table*}[t]
\setlength{\tabcolsep}{1.5pt}
\centering
\begin{tabular}{cc|cccc|cccc|cccc|cccc|cccc|cccc}
\hline
\multicolumn{2}{c|}{Data}         & \multicolumn{4}{c|}{COLMAP\cite{schoenberger2016sfm}}                & \multicolumn{4}{c|}{GLOMAP\cite{pan2024glomap}}                & \multicolumn{4}{c|}{MMA\cite{cui2022mma}}                     & \multicolumn{4}{c|}{MCSfM\cite{MCSfM}}                          & \multicolumn{4}{c|}{DMRA+MGP}                                                     & \multicolumn{4}{c}{MGSfM}                                                        \\ \hline
\multicolumn{1}{c|}{Name} & $N$   & $\tilde{e}_r$ & \multicolumn{1}{c|}{$\bar{e}_r$}  & $\tilde{e}_t$ & $\bar{e}_t$ & $\tilde{e}_r$ & \multicolumn{1}{c|}{$\bar{e}_r$}  & $\tilde{e}_t$ & $\bar{e}_t$ & $\tilde{e}_r$ & \multicolumn{1}{c|}{$\bar{e}_r$} & $\tilde{e}_t$ & $\bar{e}_t$ & $\tilde{e}_r$ & \multicolumn{1}{c|}{$\bar{e}_r$}  & $\tilde{e}_t$ & $\bar{e}_t$  & $\tilde{e}_r$ & \multicolumn{1}{c|}{$\bar{e}_r$}  & $\tilde{e}_t$ & $\bar{e}_t$  & $\tilde{e}_r$ & \multicolumn{1}{c|}{$\bar{e}_r$}  & $\tilde{e}_t$ & $\bar{e}_t$  \\ \hline
\multicolumn{1}{c|}{0000} & 9216  & {1.1}  & \multicolumn{1}{c|}{{1.5}} & 17.7           & 26.1         & \textbf{0.4}  & \multicolumn{1}{c|}{\textbf{0.7}} & 1.5           & 1.9         & 0.7           & \multicolumn{1}{c|}{1.0}         & 4.7           & 5.6         & \textbf{0.4}  & \multicolumn{1}{c|}{\textbf{0.7}} & 1.0           & 1.1          & \textbf{0.4}  & \multicolumn{1}{c|}{\textbf{0.7}} & 0.9           & 1.0          & \textbf{0.4}  & \multicolumn{1}{c|}{\textbf{0.7}} & \textbf{0.8}  & \textbf{0.9} \\
\multicolumn{1}{c|}{0002} & 11508 
& 0.8           & \multicolumn{1}{c|}{1.2} & 4.0           & 5.8       
& 0.9           & \multicolumn{1}{c|}{1.2} & 5.3           & 13.9        & 10.7          & \multicolumn{1}{c|}{13.8}        & 30.5          & 62.2        & 0.8           & \multicolumn{1}{c|}{\textbf{1.1}} & \textbf{1.1}  & 1.5          & \textbf{0.7}  & \multicolumn{1}{c|}{\textbf{1.1}} & 1.2           & \textbf{1.4} & \textbf{0.7}  & \multicolumn{1}{c|}{\textbf{1.1}} & 1.2           & \textbf{1.4} \\
\multicolumn{1}{c|}{0003} & 828   
& -           & \multicolumn{1}{c|}{-}          & -           & -        
& 0.4           & \multicolumn{1}{c|}{0.7}          & 3.5           
& 11.2        & 3.5           & \multicolumn{1}{c|}{6.3}         & 4.7           & 17.8        & 2.6           & \multicolumn{1}{c|}{3.9}          & 1.4           & 13.9         & 0.4           & \multicolumn{1}{c|}{0.6}          & \textbf{0.8}  & 1.9          & \textbf{0.3}  & \multicolumn{1}{c|}{\textbf{0.5}} & \textbf{0.8}  & \textbf{1.1} \\
\multicolumn{1}{c|}{0004} & 9272  
& 0.7           & \multicolumn{1}{c|}{1.3}          & 7.3          & 13.2        & 0.7           & \multicolumn{1}{c|}{1.2}          & 8.4          & 29.8        & 4.6           & \multicolumn{1}{c|}{11.7}        & 17.4          & 54.1        & 0.6           & \multicolumn{1}{c|}{1.0}          & 1.3           & 1.8          & 0.6           & \multicolumn{1}{c|}{1.0}          & 1.3           & 18.3         & \textbf{0.5}  & \multicolumn{1}{c|}{\textbf{0.9}} & \textbf{1.1}  & \textbf{1.4} \\
\multicolumn{1}{c|}{0005} & 5396  
& 1.6           & \multicolumn{1}{c|}{1.8}          & 6.2           & 8.3         & 0.8           & \multicolumn{1}{c|}{1.2}          & 5.5           & 31.9         & 5.6           & \multicolumn{1}{c|}{19.7}        & 31.5          & 1e2         & 0.7           & \multicolumn{1}{c|}{1.1}          & 0.9           & \textbf{1.0} & 0.5           & \multicolumn{1}{c|}{\textbf{1.0}} & \textbf{0.8}  & 1.1          & \textbf{0.4}  & \multicolumn{1}{c|}{\textbf{1.0}} & \textbf{0.8}  & 1.2          \\
\multicolumn{1}{c|}{0006} & 7760  
& 0.8           & \multicolumn{1}{c|}{1.1}          & 17.0          & 36.0         
& 1.2           & \multicolumn{1}{c|}{1.4}          & 19.4          & 2e2         & 11.4          & \multicolumn{1}{c|}{10.9}        & 39.7          & 52.2        & 1.7           & \multicolumn{1}{c|}{1.9}          & 4.0           & 5.9          & \textbf{0.7}  & \multicolumn{1}{c|}{\textbf{0.9}} & \textbf{1.5}  & 61.7         & \textbf{0.7}  & \multicolumn{1}{c|}{1.0}          & \textbf{1.5}  & \textbf{1.9} \\
\multicolumn{1}{c|}{0007} & 2720  
& 20.7           & \multicolumn{1}{c|}{26.5}          & 88.4           & 2e3         & 1.1           & \multicolumn{1}{c|}{1.2}          & 1e2           & 3e2         & 16.2          & \multicolumn{1}{c|}{13.9}        & 1e2           & 4e2         & \textbf{0.3}  & \multicolumn{1}{c|}{\textbf{0.6}} & \textbf{1.4}  & 6.0          & 1.4           & \multicolumn{1}{c|}{1.6}          & 7e2           & 1e3          & 0.6           & \multicolumn{1}{c|}{0.8}          & 2.0           & \textbf{5.5} \\
\multicolumn{1}{c|}{0009} & 11248 
& 6.3            & \multicolumn{1}{c|}{7.7} &     61.4        & 78.5          
& 0.5           & \multicolumn{1}{c|}{\textbf{0.8}} & 1.1           & 1.5         & 2.7           & \multicolumn{1}{c|}{5.1}        & 5.9          & 19.8        & \textbf{0.4}  & \multicolumn{1}{c|}{\textbf{0.8}} & \textbf{0.9}  & \textbf{1.0} & \textbf{0.4}  & \multicolumn{1}{c|}{\textbf{0.8}} & \textbf{0.9}  & \textbf{1.0} & \textbf{0.4}  & \multicolumn{1}{c|}{\textbf{0.8}} & \textbf{0.9}  & \textbf{1.0} \\
\multicolumn{1}{c|}{0010} & 7672  
& 16.0            & \multicolumn{1}{c|}{21.8}          & 4e2            & 6e2          
& 1.8           & \multicolumn{1}{c|}{1.6}          & 2e2           & 4e2         & 14.5          & \multicolumn{1}{c|}{12.1}        & 68.0          & 2e2         & \textbf{1.0}  & \multicolumn{1}{c|}{\textbf{1.1}} & 2.7           & 8.9          & \textbf{1.0}  & \multicolumn{1}{c|}{\textbf{1.1}} & 2.1           & 3e2          & 1.1           & \multicolumn{1}{c|}{1.2}          & \textbf{1.8}  & \textbf{3.4} \\ \hline
\end{tabular}
\caption{Camera pose accuracy on the KITTI-360 dataset. $N$ shows the number of images. $\tilde{e}_r$ and $\bar{e}_r$ respectively denote the median and mean rotation error in degrees; $\tilde{e}_t$ and $\bar{e}_t$ respectively denote the median and mean position error in meters. The best results are \textbf{bolded}.} 	
\label{tab:KITTI360}
\end{table*}
\begin{figure*}	
	\vspace{-0.3cm}
	\centering
	\includegraphics[width=\linewidth, trim = 0mm 50mm 0mm 45mm, clip]{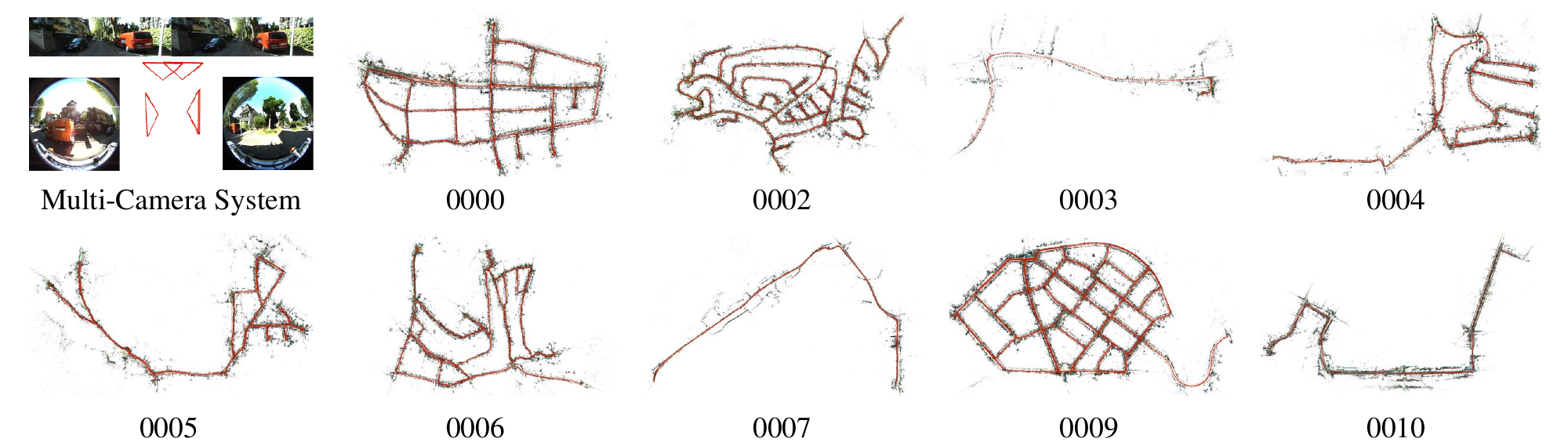}
	\caption{This figure shows the multi-camera system setup and the reconstruction results estimated by MGSfM on the KITTI-360 dataset.}
	\label{fig:KITTI360_result}
	\vspace{-0.3cm}
\end{figure*}
\begin{figure}[tp]
    \centering
    \includegraphics[width=\linewidth, trim = 0mm 0mm 0mm 5mm, clip]{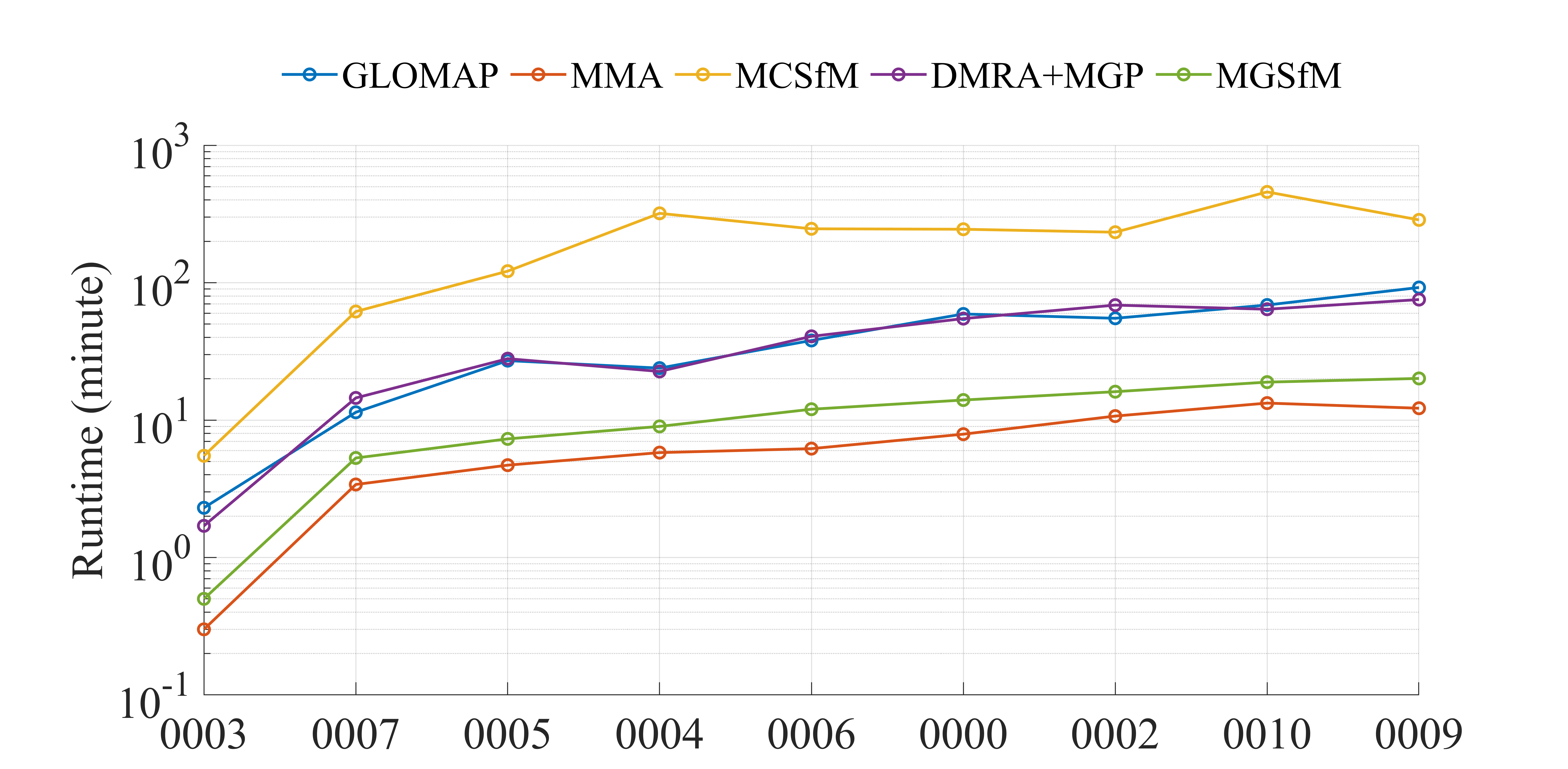}
    \caption{
    Runtime comparison (log scale) of different methods on the KITTI-360 dataset. Scenes are sorted by the ascending runtime of MGSfM to facilitate visualization.}
    \label{fig:run_time}
    \vspace{-0.5cm}
\end{figure}

\subsection{Evaluation of KITTI-360 Datasets}
The KITTI-360 dataset was captured using a station wagon equipped with a $180^\circ$ fisheye camera on each side and a $90^\circ$ perspective stereo camera at the front (see the top-left corner of \cref{fig:KITTI360_result}). To mitigate the impact of dynamic objects, we also employ the semantic segmentation algorithm from \cite{chen2017rethinking}. Compared to the KITTI Odometry benchmark, KITTI-360 is more challenging due to the wider spatial distribution of the cameras, which results in almost no overlap between the fisheye and stereo cameras. The accuracy of estimated camera poses is summarized in \cref{tab:KITTI360}, and the reconstruction results of MGSfM are shown in \cref{fig:KITTI360_result}.

Since COLMAP does not incorporate multi-camera constraints and suffers from error accumulation, its estimated camera poses exhibit significant errors. GLOMAP achieves accuracy comparable to that of MGSfM only in data 0009, where there are few outlier feature matches and sufficient loop closures. The camera rotations estimated by MMA exhibit large errors because it performs rotation averaging by jointly estimating internal camera rotations and global reference rotations, rendering it sensitive to outlier relative rotations. Although MCSfM mitigates the error accumulation by fusing multi-camera constraints, it only outperforms MGSfM in the small-scale scene 0007. 
By initializing camera positions and 3D points before the angle-based refinement stage using camera-to-point constraints, MGSfM demonstrates higher robustness and accuracy than DMRA+MGP in challenging scenes such as data 0007 and data 0010. Although DMRA+MGP and MGSfM use the same camera rotations, MGSfM achieves better results in most datasets.

The reconstruction runtimes are shown in \cref{fig:run_time}. Except for MMA, which only uses relative translations for translation averaging, MGSfM is faster than all other methods in comparison and more than $10\times$ faster than MCSfM.
Although the method DMRA+MGP fuses multi-camera constraints, its efficiency is not significantly improved as the bilinear objective requires additional optimization of numerous redundant scale variables for all feature rays.
% In conclusion, by avoiding the optimization of redundant variables during angle-based refinement, our method yields better robustness and efficiency than bilinear methods like DMRA+MGP.

\subsection{Evaluation of Self-Collected Datasets}
\begin{figure*}	
	\setlength{\abovecaptionskip}{-0.0cm}
	% \setlength{\belowcaptionskip}{-0.15cm}
	% \vspace{-0.3cm}
	\centering
	\includegraphics[width=\linewidth, trim = -4mm 50mm -4mm 50mm, clip]{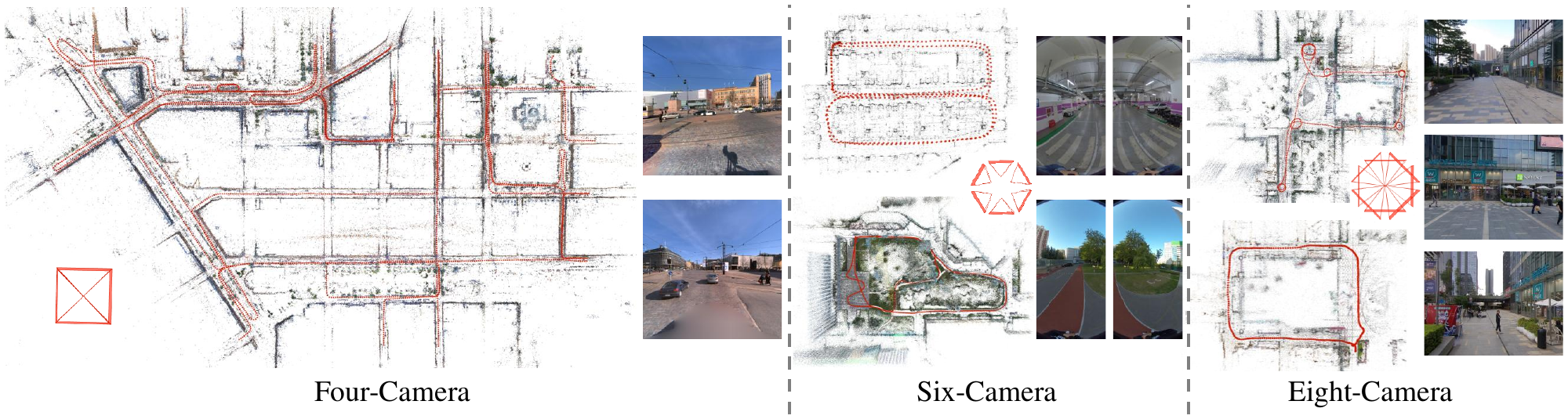}
	\caption{This figure shows the reconstruction results of five self-collected datasets. The left dataset contains 12k+ images. The middle two datasets contain 2k+ and 5k+ images, respectively. The right two datasets contain 6k+ and 2k+ images, respectively.}
	\label{fig:self}
	\vspace{-0.3cm}
\end{figure*}

To show the scalability of our system, we conduct experiments on five self-collected datasets captured by three types of multi-camera systems. The datasets and the corresponding reconstruction results from MGSfM are shown in \cref{fig:self}. Our system successfully reconstructs all scenes across various scales, demonstrating its effectiveness across different multi-camera systems. Especially for the four-camera system dataset named STREET, our system only needs half an hour to perform the motion averaging on more than 12,000 images. More results are shown in our \supp.

% 定性展示，顺便展示RA的过滤
\subsection{Ablation Study}
We use the KITTI-360 dataset~\cite{Liao2022PAMI} to analyze the impact of different input image observations and different angle-based objective functions on the performances of multi-camera translation averaging.
As shown in \cref{tab:ablation}, using the same known global rotations and view graph as input, we compare the accuracy of camera positions from six multi-camera translation averaging methods that respectively use only relative translations (denoted as ``Only trans"), only feature tracks (denoted as ``Only tracks"), or a hybrid of both (denoted as ``Hybrid"), while employing either bilinear or non-bilinear angle-based objective functions. 

For methods exclusively using relative translations, we first employ \cref{equ:distance-based MTA} to initialize camera positions, followed by angle-based refinement. As shown in the ``Only trans" column of \cref{tab:ablation}, bilinear formulations sometimes yield significantly larger mean distance errors by disrupting the continuity of the camera trajectory, whereas non-bilinear formulations are more robust to outlier relative translations. 
For methods exclusively using feature tracks, camera positions and 3D points are jointly optimized with random initialization. As shown in the ``Only tracks" column, bilinear formulations (corresponding to DMRA+MGP in \cref{tab:KITTI360}) exhibit better accuracy than their non-bilinear counterparts, indicating that the non-bilinear function is not good at handling random initializations.
For methods using hybrid input, camera positions and 3D points are jointly refined with reasonable initial values. As shown in the ``Hybrid" column, non-bilinear formulations (corresponding to MGSfM in \cref{tab:KITTI360}) show better robustness than bilinear formulations and achieve higher accuracy than methods relying solely on translations, particularly in the challenging data 0010.

In conclusion, with robust initialization, non-bilinear formulations generally yield better performance than bilinear ones, and by incorporating camera-to-point constraints from feature tracks, hybrid methods achieve improved accuracy.
\begin{table}[t]
\setlength{\tabcolsep}{1.5pt}
\centering
\scalebox{0.94}
{
\begin{tabular}{c|cccc|cccc|cccc}
\hline
Input     
& \multicolumn{4}{c|}{Only trans}                                      
& \multicolumn{4}{c|}{Only tracks}                                              
& \multicolumn{4}{c}{Hybrid}                                                    \\ \hline
{Bilinear} 
& \multicolumn{2}{c|}{yes}                    
& \multicolumn{2}{c|}{no} 
& \multicolumn{2}{c|}{yes}                    
& \multicolumn{2}{c|}{no} 
& \multicolumn{2}{c|}{yes}                    
& \multicolumn{2}{c}{no} \\ \hline
data      
& $\tilde{e}_t$ & \multicolumn{1}{c|}{$\bar{e}_t$} 
& $\tilde{e}_t$     & $\bar{e}_t$    
& $\tilde{e}_t$ & \multicolumn{1}{c|}{$\bar{e}_t$} 
& $\tilde{e}_t$    & $\bar{e}_t$    
& $\tilde{e}_t$ & \multicolumn{1}{c|}{$\bar{e}_t$} 
& $\;\;\tilde{e}_t\;$    & $\;\bar{e}_t\;\;$   \\ \hline
0000      
& 0.9           & \multicolumn{1}{c|}{1.0}         & 0.9              & 1.0            
& 0.9           & \multicolumn{1}{c|}{1.0}         & 1.1              & 2e1            
& \textbf{0.8}           & \multicolumn{1}{c|}{\textbf{0.9}}         & \textbf{0.8}              & \textbf{0.9}           \\
0002      
& 1.3           & \multicolumn{1}{c|}{2.5}         & 1.4              & 1.6            
& \textbf{1.2}           & \multicolumn{1}{c|}{\textbf{1.4}}         & 1.5              & 1e2            
& 1.3           & \multicolumn{1}{c|}{3.5}         & \textbf{1.2}              & \textbf{1.4}           \\
0003      
& 1.8           & \multicolumn{1}{c|}{5e1}        & \textbf{0.7}               & 1.2            
& 0.8           & \multicolumn{1}{c|}{1.9}         & 0.8              & 1.7            
& 0.8           & \multicolumn{1}{c|}{1.7}        & 0.8               & \textbf{1.1}           \\
0004      
& 1.2           & \multicolumn{1}{c|}{1.4}         & \textbf{1.1}              & \textbf{1.4}            
& 1.3           & \multicolumn{1}{c|}{2e1}        & 1.2              & 2e1           
& 1.7           & \multicolumn{1}{c|}{2.0}         & \textbf{1.1}              & \textbf{1.4}           \\
0005      
& 0.9           & \multicolumn{1}{c|}{1.2}         & 1.0              & 1.2            
& \textbf{0.8}           & \multicolumn{1}{c|}{\textbf{1.1}}         & 1.0              & 2e2            
& \textbf{0.8}           & \multicolumn{1}{c|}{1.2}         & \textbf{0.8}              & 1.2           \\
0006      
& 1.7           & \multicolumn{1}{c|}{2.3}         & 1.8              & 2.3            
& 1.5           & \multicolumn{1}{c|}{6e1}        & 1.9              & 1e2            
& \textbf{1.5}           & \multicolumn{1}{c|}{\textbf{1.9}}         & \textbf{1.5}              & \textbf{1.9}           \\
0007      
& 1e2           & \multicolumn{1}{c|}{3e2}         & 2.0              & 8.2            
& 7e2           & \multicolumn{1}{c|}{1e3}         & 7e2              & 9e2            
& 3.1           & \multicolumn{1}{c|}{2e2}         & \textbf{2.0}              & \textbf{5.5}           \\
0009      
& \textbf{0.9}           & \multicolumn{1}{c|}{\textbf{1.0}}         & \textbf{0.9}              & \textbf{1.0}            
& \textbf{0.9}           & \multicolumn{1}{c|}{\textbf{1.0}}         & 1.8              & 5e1           
& \textbf{0.9}           & \multicolumn{1}{c|}{\textbf{1.0}}         & \textbf{0.9}              & \textbf{1.0}           \\
0010      
& 5e1          & \multicolumn{1}{c|}{9e1}         & 2.5              & 4.7            
& 2.1          & \multicolumn{1}{c|}{3e2}          & 4e2              & 6e2            
& 5e1          & \multicolumn{1}{c|}{9e1}         & \textbf{1.8}              & \textbf{3.4}           \\ \hline
\end{tabular}
}
\caption{Camera position accuracy with different input image observations and different angle-based functions.} 	
\label{tab:ablation}
\vspace{-0.3cm}
\end{table}
% \subsection{Discussion}
% 给出KITTI360所有数据基于普通RA和中值得到相对旋转的精度
% 针对精度低且BA后结果不好的数据，采用增量方法提供的先验或给相对位姿真值的精度，来解决普通RA失败的问题

\section{Conclusion}
\label{sec:conclusion}
In this paper, we propose a multi-camera-driven global SfM pipeline that begins with decoupled multi-camera rotation averaging, followed by hybrid multi-camera translation averaging. Extensive experiments demonstrate that our approach surpasses many state-of-the-art SfM methods in terms of efficiency, accuracy, scalability, and robustness. 
% In future work, we plan to investigate a multi-camera distributed SfM framework to further enhance the performance of SfM systems for large-scale scenes.
% In the future, we plan to integrate additional information, such as coarse camera positions from GPS, to further enhance the performance of the global SfM system.

\noindent\textbf{Acknowledgments} This work was supported by the National Key R\&D Program of China (No.2023YFB3906600), the National Natural Science Foundation of China (No.U22B2055, U23A20386 and 62273345), and the Beijing Natural Science Foundation (No.L223003).

{\small
\bibliographystyle{ieeenat_fullname}
\bibliography{11_references}
}

\ifarxiv \clearpage \maketitlesupplementary \appendix \begin{table}[t]
\setlength{\tabcolsep}{1.5pt}
\centering
\setlength{\belowcaptionskip}{-0.2cm}
\scalebox{1.0}
{
\begin{tabular}{c|cccc|cccc|cccc}
\hline
Input &
  \multicolumn{4}{c|}{Only Trans} &
  \multicolumn{4}{c|}{Only Track} &
  \multicolumn{4}{c}{Hybrid} \\ \hline
Bilinear &
  \multicolumn{2}{c|}{yes} &
  \multicolumn{2}{c|}{no} &
  \multicolumn{2}{c|}{yes} &
  \multicolumn{2}{c|}{no} &
  \multicolumn{2}{c|}{yes} &
  \multicolumn{2}{c}{no} \\ \hline
data &
  $\tilde{e}_t$ &
  \multicolumn{1}{c|}{$\bar{e}_t$} &
  $\tilde{e}_t$ &
  $\bar{e}_t$ &
  $\tilde{e}_t$ &
  \multicolumn{1}{c|}{$\bar{e}_t$} &
  $\tilde{e}_t$ &
  $\bar{e}_t$ &
  $\tilde{e}_t$ &
  \multicolumn{1}{c|}{$\bar{e}_t$} &
  $\tilde{e}_t$ &
  $\bar{e}_t$ \\ \hline
00 &
  \textbf{0.5} &
  \multicolumn{1}{c|}{\textbf{0.7}} &
  \textbf{0.5} &
  0.8 &
  \textbf{0.5} &
  \multicolumn{1}{c|}{\textbf{0.7}} &
  0.6 &
  0.7 &
  0.6 &
  \multicolumn{1}{c|}{0.9} &
  \textbf{0.5} &
  \textbf{0.7} \\
01 &
  0.9 &
  \multicolumn{1}{c|}{2.7} &
  1.1 &
  1.6 &
  0.8 &
  \multicolumn{1}{c|}{31.7} &
  \textbf{0.5} &
  23.4 &
  \textbf{0.5} &
  \multicolumn{1}{c|}{1.2} &
  0.6 &
  \textbf{1.1} \\
02 &
  \textbf{0.9} &
  \multicolumn{1}{c|}{1.4} &
  \textbf{0.9} &
  1.4 &
  1.3 &
  \multicolumn{1}{c|}{1.8} &
  \textbf{0.9} &
  1.4 &
  1.0 &
  \multicolumn{1}{c|}{\textbf{1.2}} &
  \textbf{0.9} &
  1.4 \\
03 &
  \textbf{0.2} &
  \multicolumn{1}{c|}{\textbf{0.4}} &
  \textbf{0.2} &
  \textbf{0.4} &
  \textbf{0.2} &
  \multicolumn{1}{c|}{\textbf{0.4}} &
  \textbf{0.2} &
  \textbf{0.4} &
  \textbf{0.2} &
  \multicolumn{1}{c|}{\textbf{0.4}} &
  \textbf{0.2} &
  \textbf{0.4} \\
04 &
  \textbf{0.1} &
  \multicolumn{1}{c|}{\textbf{0.2}} &
  \textbf{0.1} &
  \textbf{0.2} &
  \textbf{0.1} &
  \multicolumn{1}{c|}{\textbf{0.2}} &
  \textbf{0.1} &
  \textbf{0.2} &
  \textbf{0.1} &
  \multicolumn{1}{c|}{\textbf{0.2}} &
  \textbf{0.1} &
  \textbf{0.2} \\
05 &
  \textbf{0.2} &
  \multicolumn{1}{c|}{\textbf{0.3}} &
  \textbf{0.2} &
  \textbf{0.3} &
  \textbf{0.2} &
  \multicolumn{1}{c|}{\textbf{0.3}} &
  \textbf{0.2} &
  \textbf{0.3} &
  \textbf{0.2} &
  \multicolumn{1}{c|}{\textbf{0.3}} &
  \textbf{0.2} &
  \textbf{0.3} \\
06 &
  \textbf{0.1} &
  \multicolumn{1}{c|}{\textbf{0.1}} &
  \textbf{0.1} &
  \textbf{0.1} &
  \textbf{0.1} &
  \multicolumn{1}{c|}{\textbf{0.1}} &
  \textbf{0.1} &
  \textbf{0.1} &
  \textbf{0.1} &
  \multicolumn{1}{c|}{\textbf{0.1}} &
  \textbf{0.1} &
  \textbf{0.1} \\
07 &
  \textbf{0.2} &
  \multicolumn{1}{c|}{\textbf{0.3}} &
  \textbf{0.2} &
  \textbf{0.3} &
  \textbf{0.2} &
  \multicolumn{1}{c|}{\textbf{0.3}} &
  \textbf{0.2} &
  \textbf{0.3} &
  \textbf{0.2} &
  \multicolumn{1}{c|}{\textbf{0.3}} &
  \textbf{0.2} &
  \textbf{0.3} \\
08 &
  1.1 &
  \multicolumn{1}{c|}{\textbf{1.4}} &
  \textbf{0.9} &
  1.5 &
  1.0 &
  \multicolumn{1}{c|}{1.5} &
  1.0 &
  \textbf{1.4} &
  \textbf{0.9} &
  \multicolumn{1}{c|}{1.6} &
  \textbf{0.9} &
  \textbf{1.4} \\
09 &
  \textbf{0.5} &
  \multicolumn{1}{c|}{1.1} &
  \textbf{0.5} &
  \textbf{1.0} &
  \textbf{0.5} &
  \multicolumn{1}{c|}{\textbf{1.0}} &
  \textbf{0.5} &
  \textbf{1.0} &
  \textbf{0.5} &
  \multicolumn{1}{c|}{\textbf{1.0}} &
  \textbf{0.5} &
  \textbf{1.0} \\
10 &
  \textbf{0.6} &
  \multicolumn{1}{c|}{\textbf{1.3}} &
  \textbf{0.6} &
  \textbf{1.3} &
  \textbf{0.6} &
  \multicolumn{1}{c|}{1.4} &
  \textbf{0.6} &
  1.4 &
  \textbf{0.6} &
  \multicolumn{1}{c|}{\textbf{1.3}} &
  \textbf{0.6} &
  \textbf{1.3} \\ \hline
\end{tabular}
}
\caption{Comparison of camera position accuracy estimated by six methods with different input image observations and different angle-based functions on KITTI Odometry dataset.} 	
\label{tab:ablation_odo}
\vspace{-0.1cm}
\end{table}
\section{Ablation Study on KITTI Odometry}
We analyze the impact of different input image observations and angle-based objective functions on the performance of multi-camera translation averaging on the KITTI Odometry benchmark~\cite{geiger2013vision}, as shown in \cref{tab:ablation_odo}. For most data, all six methods yield comparable accuracy in camera positions. Since there is sufficient overlap in the field of view between the stereo cameras, four overlapping image pairs are typically formulated between two adjacent rigid units. Consequently, the relative scales between the rigid units can be accurately estimated using only relative translations, even in cases of collinear camera motion trajectories. In data 01, however, there are fewer feature points and a higher proportion of outliers between image pairs, causing methods that rely solely on feature tracks with random initialization to easily converge to incorrect local optima. In contrast, relative translations exhibit a higher inlier ratio, as they are robustly estimated from multiple feature matches. 
Given a reasonable initial solution for camera positions and 3D points, the unbiased angle-based refinement formulated from camera-to-point constraints exhibits higher robustness to outlier feature tracks and achieves higher accuracy.
\begin{figure}[ht]
	\setlength{\belowcaptionskip}{-0.2cm}
    \centering
    \includegraphics[width=\linewidth, trim = 0mm 25mm 0mm 20mm, clip]{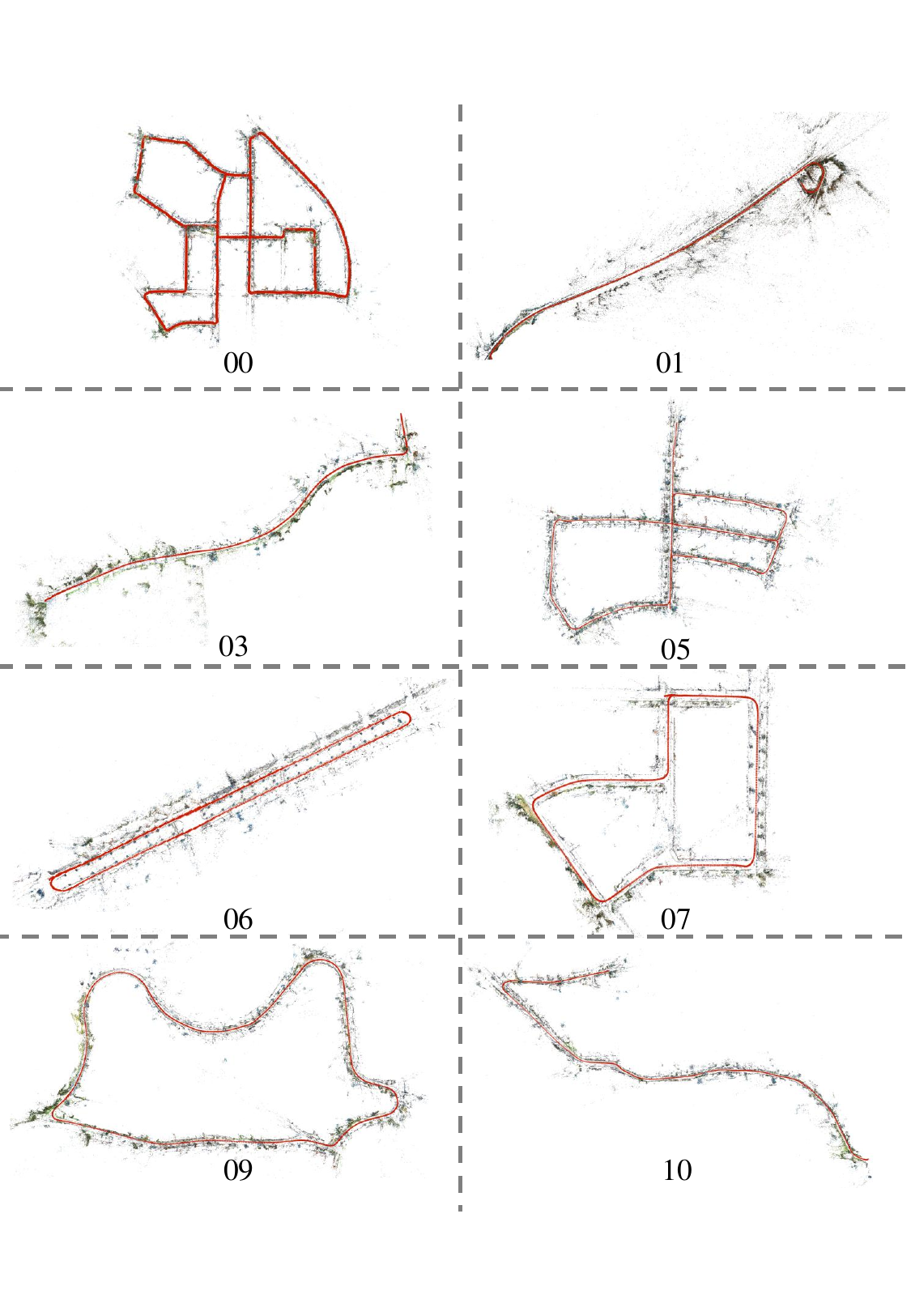}
    \caption{Our reconstruction results from the partial KITTI Odometry benchmark.}
    \label{fig:kitti_partial}
    \vspace{-0.1cm}
\end{figure}
\begin{figure}[ht]
	\setlength{\belowcaptionskip}{-0.2cm}
    \centering
    \includegraphics[width=\linewidth, trim = 0mm 0mm 0mm 0mm, clip]{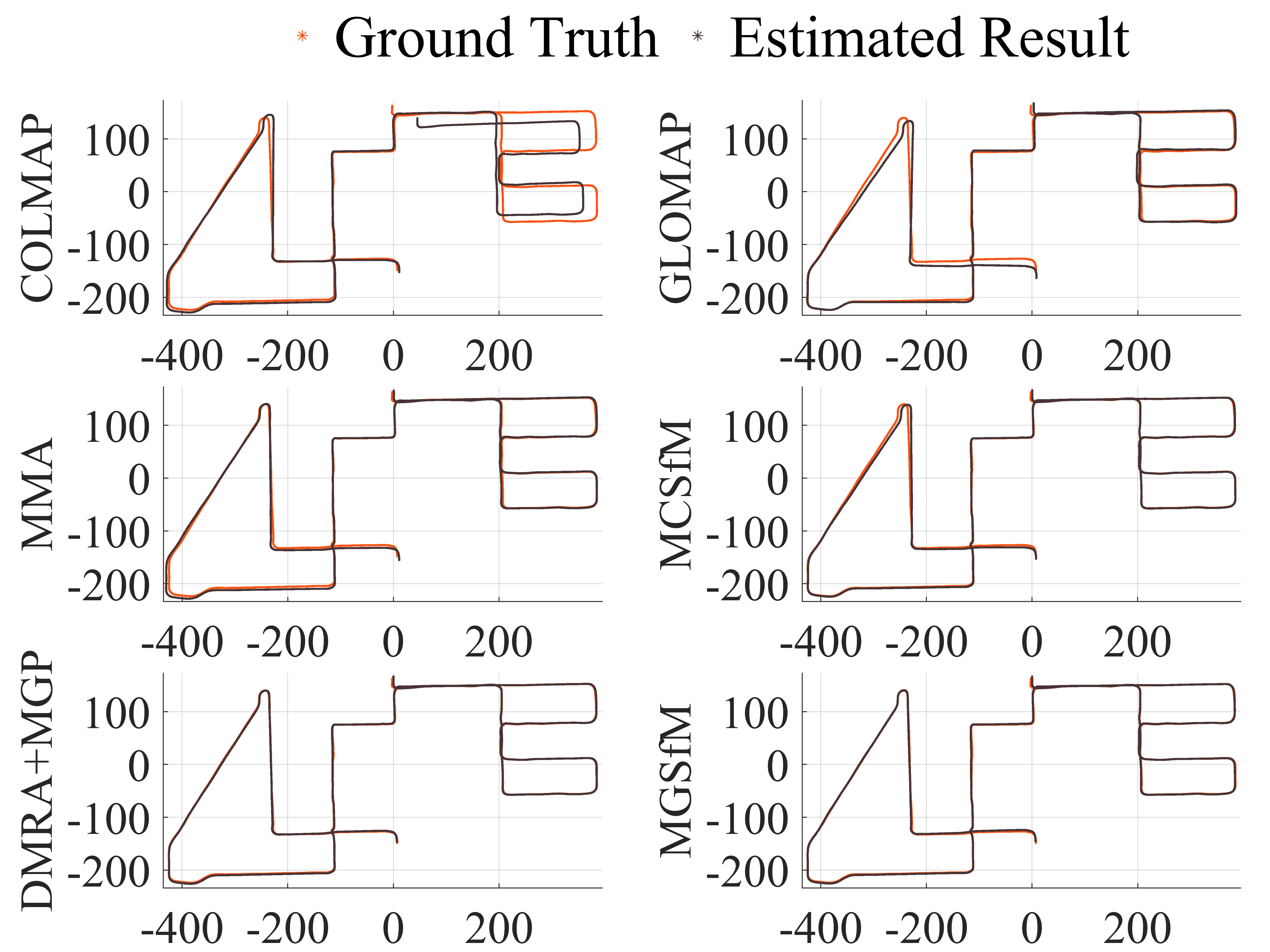}
    \caption{Comparison of camera motion trajectories on data 08 in KITTI Odometry benchmark~\cite{geiger2013vision}. The sample state-of-the-art SfM methods include COLMAP~\cite{schoenberger2016sfm}, GLOMAP~\cite{pan2024glomap}, MMA~\cite{cui2022mma}, MCSfM~\cite{MCSfM} and our proposed DMRA+MGP and MGSfM.}
    \label{fig:kitti08}
    \vspace{-0.4cm}
\end{figure}

\section{Qualitative Results on KITTI Odometry}
% In this section, we present qualitative results from the KITTI Odometry benchmark. 
\cref{fig:kitti_partial} presents additional reconstruction outcomes of MGSfM on the KITTI Odometry benchmark~\cite{geiger2013vision}. We compare the camera motion trajectories estimated by six state-of-the-art methods on the challenging large-scale sequence 08, which lacks complete loop closure. As depicted in \cref{fig:kitti08}, the red trajectory represents the ground truth camera positions, while the black trajectory represents the estimated camera positions obtained by the different methods. For incremental methods, compared to COLMAP, the incorporation of multi-camera constraints in MCSfM yields a camera trajectory that is noticeably closer to the ground truth, demonstrating the significance of multi-camera constraints in mitigating scale drift. Although GLOMAP jointly estimates all camera positions using camera-to-point constraints to achieve a uniform error distribution, its estimated camera trajectory still exhibits large errors relative to the ground truth. In contrast, MMA estimates camera positions using only relative translations but achieves higher accuracy than GLOMAP by fusing multi-camera constraints. Furthermore, global methods, such as MGSfM, that employ both camera-to-point and multi-camera constraints to jointly estimate all camera positions achieve a more uniform error distribution, resulting in trajectories that are closer to the ground truth than those produced by incremental methods such as MCSfM.

\begin{table}[t]
\setlength{\belowcaptionskip}{-0.5cm}
\setlength{\tabcolsep}{1.0pt}
\centering
\scalebox{0.9}
{
\begin{tabular}{c|cccccc|ccc|cccccc}
\hline
     & \multicolumn{6}{c|}{MCSfM~\cite{MCSfM}}                             & \multicolumn{3}{c|}{Median-RA} & \multicolumn{6}{c}{MGSfM}                              \\ \hline
Data 
& $e_r^{r1}$ & $e_r^{r2}$ & \multicolumn{1}{c|}{$e_r^{r3}$} 
& $e_t^{r1}$ & $e_t^{r2}$ & $e_t^{r3}$ 
& $e_r^{r1}$ & $e_r^{r2}$ & $e_r^{r3}$ 
& $e_r^{r1}$ & $e_r^{r2}$ & \multicolumn{1}{c|}{$e_r^{r3}$} 
& $e_t^{r1}$ & $e_t^{r1}$ & $e_t^{r3}$ \\  \hline
0000 & 1.1 & 0.5 & \multicolumn{1}{c|}{0.6} & 0.4 & 0.4 & 1.0 & 1.1      & 0.6      & 0.5      & 1.1 & 0.6 & \multicolumn{1}{c|}{0.5} & 0.7 & 0.6 & 0.9 \\
0002 & 1.1 & 0.5 & \multicolumn{1}{c|}{0.6} & 0.4 & 0.4 & 1.0 & 1.1      & 0.6      & 0.5      & 1.1 & 0.6 & \multicolumn{1}{c|}{0.5} & 0.6 & 0.3 & 1.2 \\
0003 & 1.1 & 0.5 & \multicolumn{1}{c|}{0.7} & 0.4 & 0.6 & 0.9 & 1.1      & 0.5      & 0.5      & 1.1 & 0.6 & \multicolumn{1}{c|}{0.5} & 0.6 & 0.1 & 1.3 \\
0004 & 1.1 & 0.5 & \multicolumn{1}{c|}{0.6} & 0.4 & 0.4 & 1.0 & 1.1      & 0.6      & 0.6      & 1.1 & 0.6 & \multicolumn{1}{c|}{0.5} & 0.5 & 0.3 & 1.2 \\
0005 & 1.1 & 0.5 & \multicolumn{1}{c|}{0.6} & 0.4 & 0.4 & 0.9 & 1.1      & 0.6      & 0.5      & 1.1 & 0.6 & \multicolumn{1}{c|}{0.5} & 0.6 & 0.4 & 1.1 \\
0006 & 1.1 & 0.5 & \multicolumn{1}{c|}{0.6} & 0.4 & 0.4 & 1.0 & 1.1      & 0.6      & 0.5      & 1.1 & 0.6 & \multicolumn{1}{c|}{0.5} & 0.6 & 0.2 & 1.2 \\
0007 & 1.1 & 0.5 & \multicolumn{1}{c|}{0.6} & 0.5 & 0.5 & 1.1 & 1.1      & 0.5      & 0.5      & 1.2 & 0.6 & \multicolumn{1}{c|}{0.5} & 0.6 & 0.5 & 1.2 \\
0009 & 1.1 & 0.5 & \multicolumn{1}{c|}{0.6} & 0.4 & 0.4 & 1.0 & 1.2      & 0.6      & 0.5      & 1.2 & 0.7 & \multicolumn{1}{c|}{0.5} & 0.8 & 0.5 & 0.8 \\
0010 & 1.1 & 0.5 & \multicolumn{1}{c|}{0.7} & 0.4 & 0.6 & 1.0 & 1.2      & 0.6      & 0.5      & 1.2 & 0.6 & \multicolumn{1}{c|}{0.6} & 0.7 & 0.3 & 0.8 \\ \hline
\end{tabular}
}
\caption{Accuracy of internal camera poses estimated by MCSfM and MGSfM, as well as the median of internal rotations estimated via single-camera rotation averaging, on KITTI-360 benchmark. The left camera of the stereo pair in multi-camera system is selected as the reference. 
Here, $e_r^r$ and $e_t^r$ denote the angular errors of relative rotations and relative translations, respectively, in degrees.} 	
\label{tab:rel_pose}
\end{table}

\section{Internal Pose Estimation on KITTI-360}
\label{sec:internal rotation estimation}
In this section, we compare the internal camera poses estimated by MCSfM~\cite{MCSfM} and MGSfM, as well as the median of internal camera rotations (denoted as ``Median-RA"), on the KITTI-360 benchmark~\cite{Liao2022PAMI}. As shown in \cref{tab:rel_pose}, the accuracy of the initial internal camera rotations from Median-RA is comparable to that achieved by MCSfM or to the results refined via BA in MGSfM, demonstrating the robustness of our decoupled rotation averaging method. Moreover, the accuracy of internal camera translations estimated by MCSfM and MGSfM is also comparable, indicating that MGSfM exhibits robustness on par with incremental methods.

\section{Runtime of Ablation Study on KITTI-360}
We report the runtime of six methods evaluated in an ablation study on the KITTI-360 dataset~\cite{Liao2022PAMI}. These methods respectively utilize only relative translations (denoted as ``Only trans"), only feature tracks (denoted as ``Only tracks"), or a hybrid of both (denoted as ``Hybrid"), and employ either bilinear or non-bilinear angle-based objective functions. As shown in \cref{fig:run_time_abl}, except for methods with ``Only Trans", MGSfM (Hybrid-Non-Bilinear) is faster than all other approaches. Notably, by providing a reasonable initialization of camera positions and 3D points, MGSfM achieves significantly higher efficiency than the ``Only-Track-Non-Bilinear" method, underscoring the importance of initialization.

\section{Qualitative Results on KITTI-360}
We compare the reconstruction results from several state-of-the-art SfM methods on KITTI-360 benchmark~\cite{Liao2022PAMI}, including COLMAP~\cite{schoenberger2016sfm}, GLOMAP~\cite{pan2024glomap}, MMA~\cite{cui2022mma}, MCSfM~\cite{MCSfM} and our proposed MGSfM, as shown in~\cref{fig:360_3}.

\begin{figure}[tp]
\setlength{\belowcaptionskip}{-0.3cm}
\centering
\includegraphics[width=\linewidth, trim = 0mm 0mm 0mm 10mm, clip]{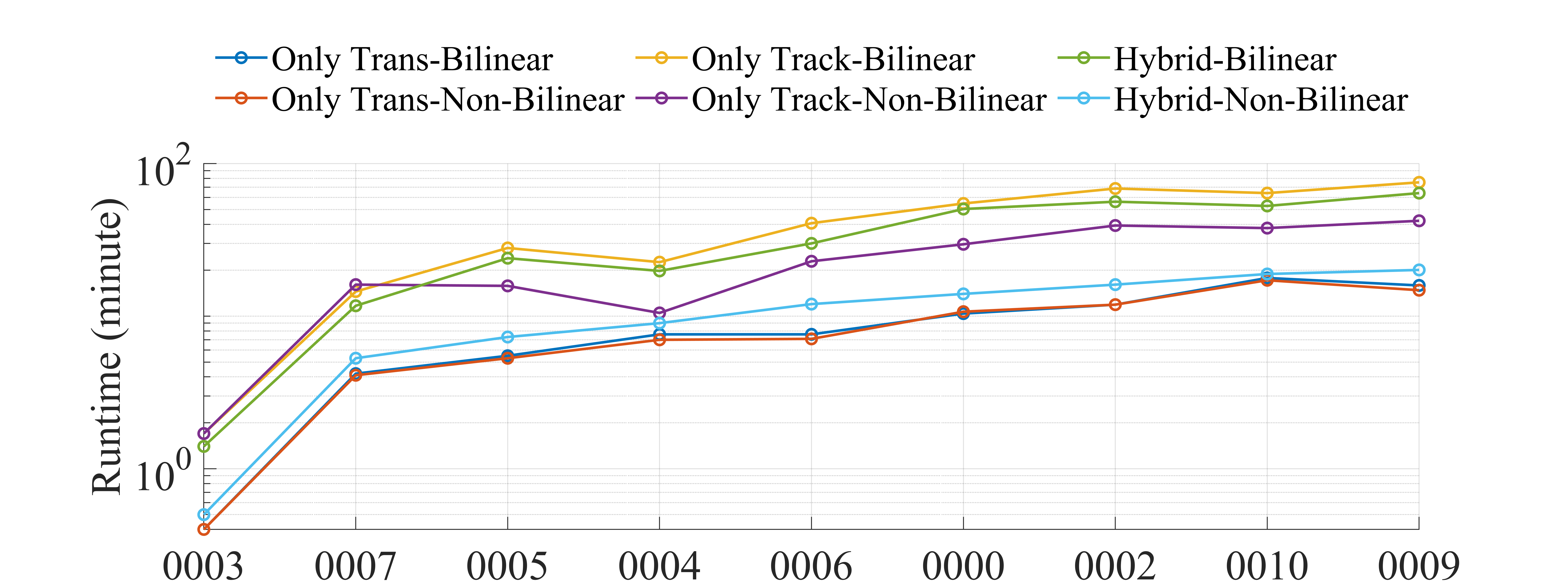}
\caption{
Runtime comparison (log scale) of six ablation study methods on the KITTI-360 dataset. Scenes are sorted by the ascending runtime of Hybrid-Non-Bilinear to facilitate visualization.}
\label{fig:run_time_abl}
\end{figure}

\section{Test on Indoor ETH3D-SLAM.}
As shown in table below, MGSfM achieves the best accuracy and efficiency ($T$ in seconds) of translation averaging.
\begin{table}[H]
\setlength{\tabcolsep}{4pt}
\centering
\scalebox{0.7}
{
\begin{tabular}{cc|ccc|ccc|ccc}
\hline
\multicolumn{2}{c|}{Data}                  & \multicolumn{3}{c|}{GLOMAP}             & \multicolumn{3}{c|}{DMRA+MGP}          & \multicolumn{3}{c}{MGSfM}              \\ \hline
\multicolumn{1}{c|}{\multirow{2}{*}{Name}} &
  \multirow{2}{*}{$N$} &
  \multicolumn{2}{c|}{AUC@} &
  \multirow{2}{*}{$T$} &
  \multicolumn{2}{c|}{AUC@} &
  \multirow{2}{*}{$T$} &
  \multicolumn{2}{c|}{AUC@} &
  \multirow{2}{*}{$T$} \\ \cline{3-4} \cline{6-7} \cline{9-10}
\multicolumn{1}{c|}{}               &      & 0.1m & \multicolumn{1}{c|}{0.5m} &      & 0.1m & \multicolumn{1}{c|}{0.5m} &     & 0.1m & \multicolumn{1}{c|}{0.5m} &     \\ \hline
\multicolumn{1}{c|}{ceiling\_1}     & 3190 & 18.5 & \multicolumn{1}{c|}{66.2} & 240  & 32.8 & \multicolumn{1}{c|}{78.0} & 154 & \textbf{59.7} & \multicolumn{1}{c|}{\textbf{87.4}} & \textbf{34}  \\
\multicolumn{1}{c|}{desk\_3}        & 4132 & 86.4 & \multicolumn{1}{c|}{97.3} & 462  & 95.3 & \multicolumn{1}{c|}{99.1} & 467 & \textbf{95.8} & \multicolumn{1}{c|}{\textbf{99.2}} & \textbf{89}  \\
\multicolumn{1}{c|}{large\_loop\_1} & 3022 & 70.0 & \multicolumn{1}{c|}{92.7} & 250  & 81.1 & \multicolumn{1}{c|}{96.0} & 166 & \textbf{87.9} & \multicolumn{1}{c|}{\textbf{97.6}} & \textbf{30}  \\
\multicolumn{1}{c|}{motion\_1}      & 4732 & 25.7 & \multicolumn{1}{c|}{70.3} & 885  & 22.4 & \multicolumn{1}{c|}{70.3} & 678 & \textbf{46.5} & \multicolumn{1}{c|}{\textbf{87.6}} & \textbf{158} \\
\multicolumn{1}{c|}{reflective\_1}  & 9202 & 79.1 & \multicolumn{1}{c|}{95.7} & 3239 & 87.5 & \multicolumn{1}{c|}{96.3} & 670 & \textbf{91.3} & \multicolumn{1}{c|}{\textbf{97.2}} & \textbf{335} \\
\multicolumn{1}{c|}{repetitive}     & 3932 & 66.9 & \multicolumn{1}{c|}{93.2} & 90   & 83.0 & \multicolumn{1}{c|}{96.2} & 104 & \textbf{91.4} & \multicolumn{1}{c|}{\textbf{97.8}} & \textbf{23}  \\ \hline
\end{tabular}
}
% \caption{ } 	
\label{tab:eth3d_slam}
\end{table}

\section{Results on Self-Collected Datasets}
We compare our method with several state-of-the-art SfM methods, including COLMAP~\cite{schoenberger2016sfm}, GLOMAP~\cite{pan2024glomap}, and MCSfM~\cite{MCSfM}, on self-collected datasets. The CAMPUS dataset comprises more than 29,000 images covering an area of approximately 520,000 $m^2$. Qualitative results for the CAMPUS dataset are presented in \cref{fig:ucas_all}. The runtime of MGSfM is 66 minutes, compared to approximately 1588 minutes for COLMAP, 580 minutes for GLOMAP, 401 minutes for MCSfM and 51 minutes for MMA. Due to the lack of multi-camera constraints in COLMAP and GLOMAP, both methods are sensitive to outlier feature matches. Although MCSfM incorporates multi-camera constraints, its inherent error accumulation limits its accuracy.

The STREET dataset, which is captured by a four-camera system, comprises more than 12,000 images covering an area of approximately 500,000 $m^2$. Qualitative results for the STREET dataset are shown in \cref{fig:SV1}. Only MGSfM accurately reconstructs the tracks of the roads.
\begin{figure*}[p]	
	% \setlength{\abovecaptionskip}{-0.0cm}
	% \setlength{\belowcaptionskip}{-0.15cm}
	% \vspace{-0.3cm}
	\centering
	\includegraphics[width=0.83\linewidth, trim = 0mm 0mm 0mm 0mm, clip]{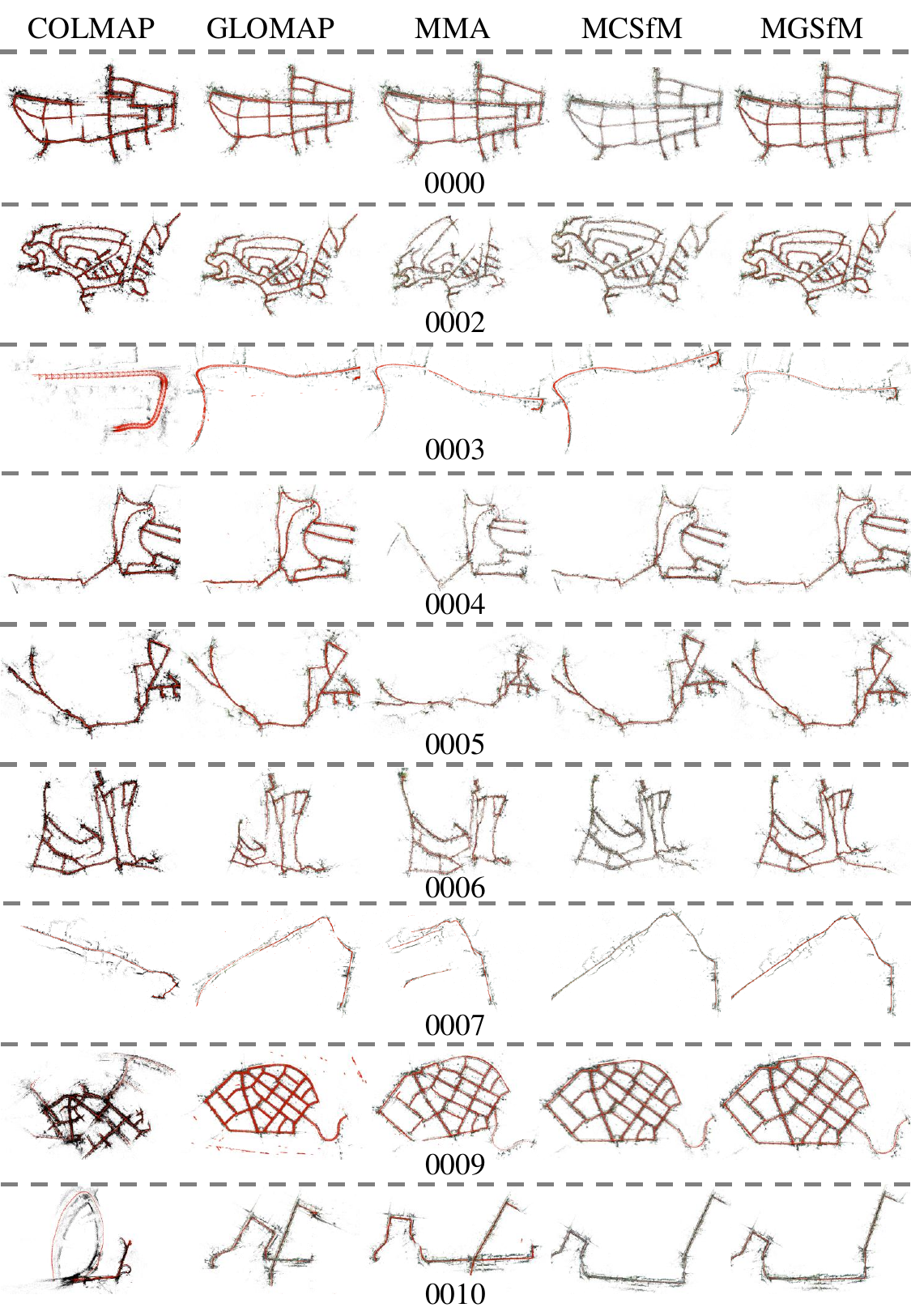}
	\caption{Comparison of reconstruction results on KITTI-360 benchmark~\cite{Liao2022PAMI}. From the qualitative comparison, our method demonstrates superior robustness compared to COLMAP~\cite{schoenberger2016sfm}, GLOMAP~\cite{pan2024glomap} and MMA~\cite{cui2022mma}. As highlighted in our main manuscript, our system achieves a computational efficiency approximately 10 times faster than that of  MCSfM~\cite{MCSfM}.} 
	\label{fig:360_3}
\end{figure*}
\begin{figure*}[p]	
	% \setlength{\abovecaptionskip}{-0.0cm}
	% \setlength{\belowcaptionskip}{-0.15cm}
	% \vspace{-0.3cm}
	\centering
	\includegraphics[width=0.83\linewidth, trim = 0mm 0mm 0mm 0mm, clip]{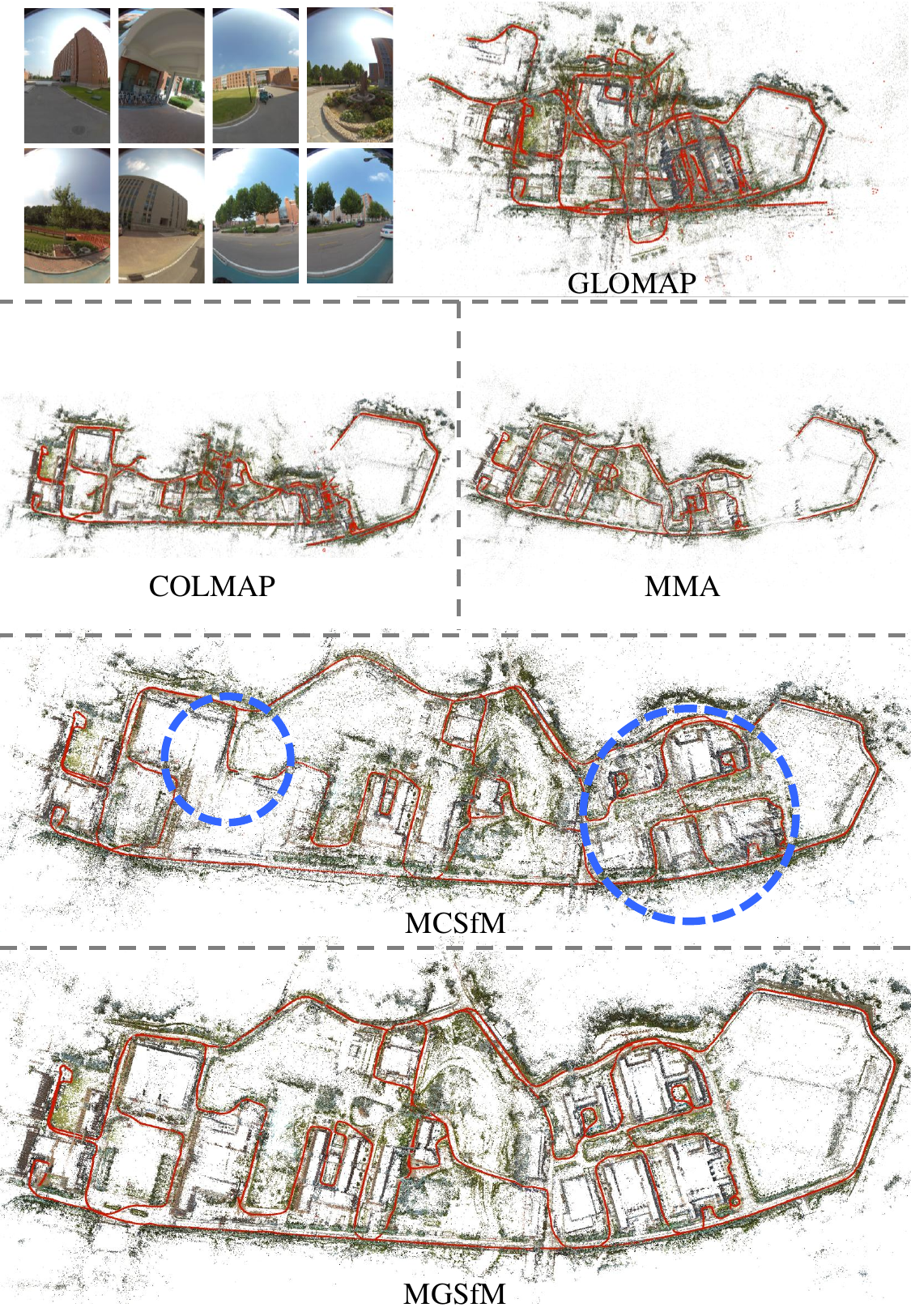}
	\caption{Comparison of reconstruction results on the self-collected CAMPUS dataset. The state-of-the-art SfM methods compared include COLMAP~\cite{schoenberger2016sfm}, GLOMAP~\cite{pan2024glomap}, MCSfM~\cite{MCSfM}, MMA~\cite{cui2022mma} and our proposed MGSfM. 
    The results produced by COLMAP, GLOMAP and MMA are wrong.
    For the result produced by MCSfM,
    the area enclosed by the blue circle indicates an incorrect reconstruction structure.}
	\label{fig:ucas_all}
\end{figure*}
\begin{figure*}[p]	
	% \setlength{\abovecaptionskip}{-0.0cm}
	% \setlength{\belowcaptionskip}{-0.15cm}
	% \vspace{-0.3cm}
	\centering
	\includegraphics[width=0.83\linewidth, trim = 0mm 0mm 0mm 0mm, clip]{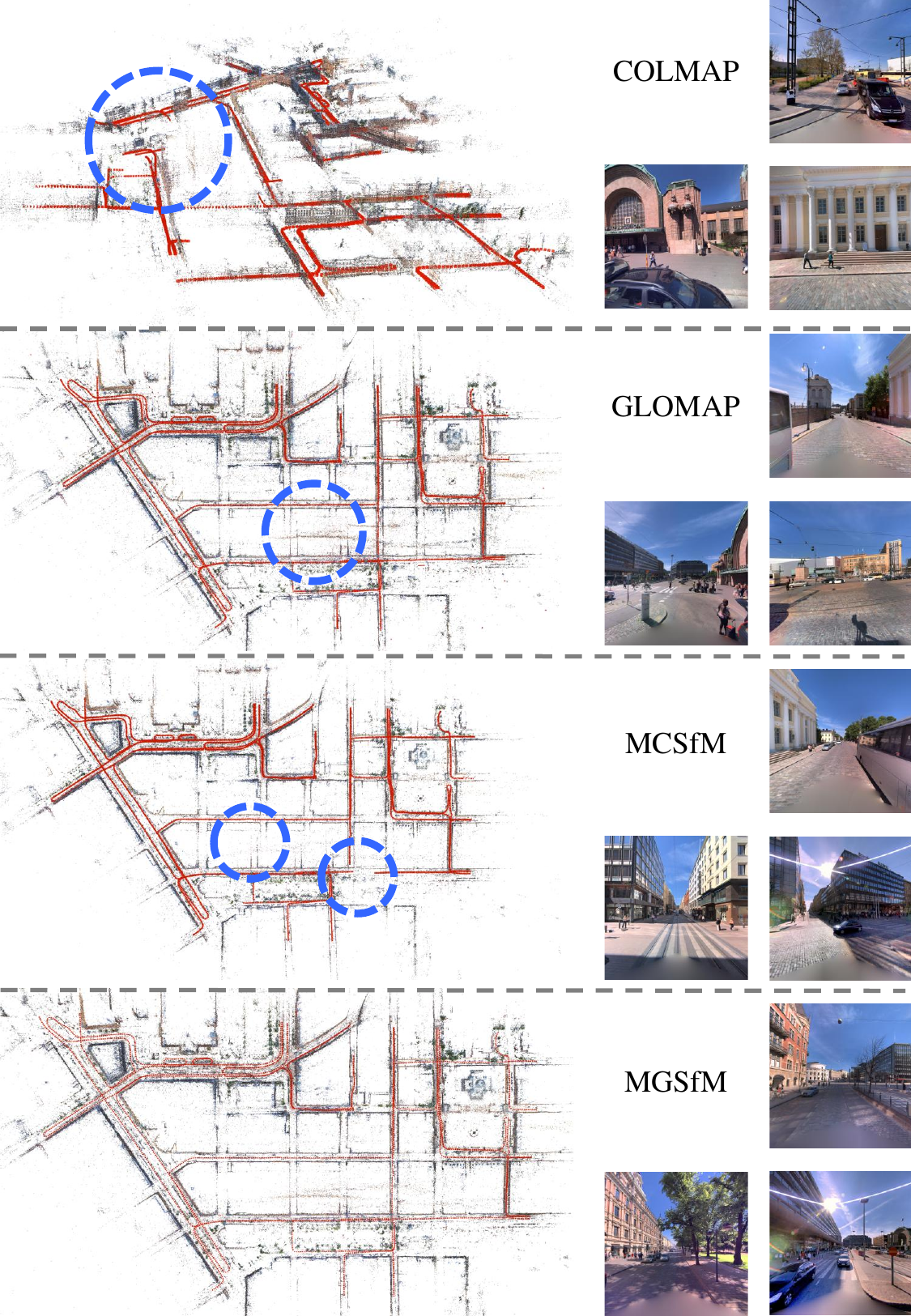}
	\caption{Comparison of reconstruction results on the self-collected STREET dataset. Some sample images are shown near the name of the compared method. The state-of-the-art SfM methods compared include COLMAP~\cite{schoenberger2016sfm}, GLOMAP~\cite{pan2024glomap}, MCSfM~\cite{MCSfM}, MMA~\cite{cui2022mma}, and our proposed MGSfM. The area enclosed by the blue circle indicates an incorrect scene structure in the reconstruction.
}
	\label{fig:SV1}
\end{figure*}
 \fi

\end{document}